\documentclass[journal]{IEEEtran}

\usepackage{graphicx}
\usepackage{amsmath,amssymb,amsfonts}
\usepackage{algorithm}
\usepackage{algpseudocode}
\usepackage{array}
\usepackage{tabularx}
\usepackage{textcomp}
\usepackage{xcolor}
\usepackage{pifont} 
\newcommand{\cmark}{\ding{51}}
\newcommand{\xmark}{\ding{55}}
\newcommand{\tablehead}[1]{\begingroup\renewcommand{\arraystretch}{0.85}\begin{tabular}[c]{@{}c@{}}#1\end{tabular}\endgroup}
\usepackage[numbers,sort&compress]{natbib}
\usepackage[colorlinks=true,citecolor=blue,linkcolor=blue,urlcolor=blue]{hyperref}
\usepackage{cuted} 
\usepackage{dblfloatfix} 
\graphicspath{{images/}}

\begin{document}

\title{Light-Loco-Parkour: Versatile Perceptive Whole-Body Locomotion via Multi-Skill Distillation}

\author{Hongming~Chen,
        Zhuoran~Li,
        Hongxi~Wang,
        Jiangpeng~Hu,
        Ziliang~Li,
        Peize~Liu,
        QingRui~Zhao, \\
        Xuhao~Liu,
        Liang~Pan,
        Ximin~Lyu,
        Yuntao~Ma$^\dagger$,
        and~Tingxiang~Fan$^\dagger$\\
        Light Origins
\thanks{$^\dagger$Light Origins robotics team Co-Lead}}

\markboth{}%
{Chen \MakeLowercase{\textit{et al.}}: Light-Loco-Parkour: Perceptive Whole-Body Locomotion via Multi-Skill Distillation}

\maketitle

\begin{strip}
  \vspace*{-3.0em}\vspace*{-1cm}
  \centering
  \includegraphics[width=\textwidth]{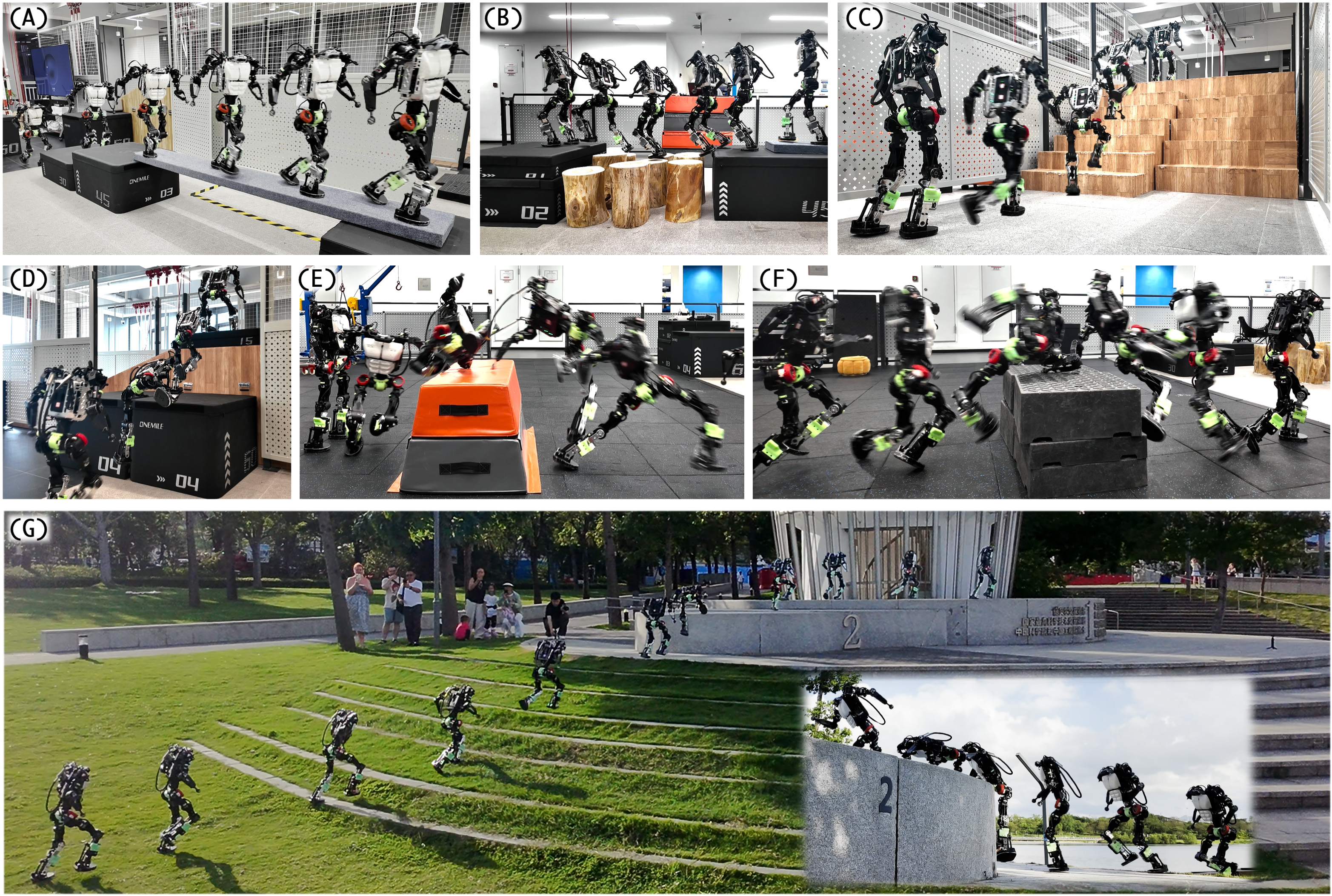}\\[3pt]
  \refstepcounter{figure}\label{fig:teaser}%
  \parbox{\textwidth}{\footnotesize
    Fig.~\thefigure.\quad\textbf{Light-Loco-Parkour} deployed on Lightbot~0, a custom-built $90$~cm humanoid, and performs perceptive whole-body locomotion across indoor and outdoor terrains with onboard sensors and compute.
    \textbf{(A)} Crossing a narrow plank bridge, where precise foot placement is essential.
    \textbf{(B)} Traversing stepping stones over sparse footholds.
    \textbf{(C)} Climbing a staircase.
    \textbf{(D)} A double climb onto a tall platform beyond leg reach, bracing with the arms across two stages.
    \textbf{(E)} A speed-vault over a box, planting a hand and swinging the legs across.
    \textbf{(F)} A reverse-vault that clears a box back-first.
    \textbf{(G)} Outdoor experiments across natural terrain.
    Videos are available at \href{https://light-loco-parkour.github.io/}{\textcolor{black}{\textbf{https://light-loco-parkour.github.io/}}}.}
\end{strip}

\begin{abstract}
Existing humanoid whole-body control systems still fall short of the way humans move through cluttered terrain: they either track expressive whole-body references without terrain generalization, or react to terrain online while leaving the arms, torso, and knees largely unused.
We present \texttt{Light-Loco-Parkour} (LightLP), an end-to-end perceptive whole-body locomotion system that closes this gap with a single deployable policy.
Conditioned only on onboard depth and a velocity command, the policy decides when to walk, balance, climb, step down, or vault, with no reference input, skill label, hand-coded gate, or runtime motion graph.
Compared with prior humanoid systems, LightLP makes three contributions.
First, it introduces a whole-body perceptive-control pipeline that extends an RL-trained, velocity-tracking locomotion policy with parkour skills learned from object-interacting motions, so the same policy tracks velocity in open terrain, executes whole-body traversal at obstacles, and resumes locomotion afterward.
Second, it acquires terrain-conditioned skills from sparse seeds by expanding a single motion into dynamically feasible, terrain-paired references across obstacle geometry, rather than relying on a large motion corpus.
Third, it learns autonomous skill transitions from reward, letting the policy decide when and which whole-body skill to invoke from depth and command alone, with no one-hot skill label, hand-coded state machine, or runtime motion generator.
Simulation and real-world experiments show high success across both benchmarked terrains and unseen obstacle variations, and the same policy transfers zero-shot to indoor and outdoor hardware experiments.
These results demonstrate autonomous perceptive whole-body locomotion on a humanoid in outdoor settings, using only onboard sensing and a single deployable policy.
\end{abstract}


\section{Introduction}

\IEEEPARstart{H}{umanoid} robots are designed to operate in the cluttered, human-centric environments that wheeled and tracked machines cannot reach.
Moving through such spaces is rarely a matter of the legs alone: traversing an obstacle often demands the \emph{entire} body, with the hands vaulting, the knees bracing, and the torso leaning in continuous coordination with what the robot perceives.
A humanoid that can sense the terrain ahead and recruit its whole body to negotiate it would unlock applications ranging from search-and-rescue over rubble, to inspection of cluttered industrial sites, to operation in the everyday human surroundings these machines are ultimately built for.

Recent learned controllers have demonstrated several humanoid capabilities: they can track whole-body reference motions~\citep{peng2018deepmimic,liao2025beyondmimic}, walk on real hardware~\citep{radosavovic2024real,gu2024advancing}, and support teleoperated whole-body tasks~\citep{darvish2023teleoperation}.
Most demonstrated capabilities, however, remain limited to flat or mildly uneven terrain.
Traversal of rough, obstacle-rich environments remains less common, especially when success requires non-foot contacts.
Progress toward this setting has mainly followed two learning paradigms: motion imitation and locomotion.

Mimic-based methods learn from a human-motion prior.
A reference trajectory, retargeted from motion capture or video, is provided to the policy as part of its observation, and the policy is rewarded for tracking it~\citep{peng2018deepmimic,peng2021amp,margolis2025softmimic}.
This formulation has been used to reproduce dynamic whole-body skills, including cartwheels and backflips~\citep{peng2018deepmimic}, and recent systems have scaled reference tracking to larger and more diverse motion sets~\citep{wang2025hil,liao2025beyondmimic,luo2025sonic,luo2023perpetual}.
Its limitation is that the deployed behavior remains tied to the reference distribution.
When the robot or environment deviates from the captured motion, the policy may enter out-of-distribution (OOD) states.
Moreover, interaction data with objects and terrain is scarce, and without exteroception the policy cannot adapt its motion to scene geometry that differs from the demonstration~\citep{yang2025omniretarget}.

Locomotion methods take a different approach.
They optimize a policy directly with reinforcement learning to track a commanded planar velocity $(v_x, v_y, \omega_z)$.
Without a reference trajectory, these policies can discover robust gaits for stairs, slopes, and unstructured terrain, and have transferred successfully to real robots~\citep{zhuang2023robot,rudin2025parkour,kim2025high,zhuang2026deep}.
However, reward-only learning has difficulty discovering contact-rich whole-body skills on humanoids.
In practice, policies often converge to leg-dominant behaviors, while coordinated use of the arms, torso, and knees requires extensive reward shaping and does not scale easily across diverse parkour skills.

These complementary limitations motivate whole-body locomotion~\citep{wu2026perceptive,zhang2026learning}, which combines reference-driven whole-body motion with terrain-conditioned perceptive control.
Perception can condition a skill on the observed terrain, while whole-body contact extends locomotion beyond foot-only traversal.
Existing approaches, however, still inherit important costs from reference-based learning, including offline motion generation and motion-capture datasets that scale poorly with the number of skills, contacts, and terrain variations.

Realizing whole-body locomotion that is at once capable, generalizable, and deployable raises three challenges.
1) The first is data scarcity.
Whole-body parkour requires reference motions that are not only expressive, but also precisely paired with the terrain contacts that make the motion possible.
Motion-capture clips must be retargeted across different embodiments and then aligned to objects, a process that easily produces floating contacts, scene penetration, or dynamically infeasible references; teleoperation cannot provide this data either, since current humanoids and teleoperation interfaces cannot reliably execute forceful whole-body parkour interactions in the first place.
2) The second challenge is learning a \emph{generalizable} whole-body skill from such references.
Most reference-based skills are demonstrated in a specific environment, with fixed object size, shape, and placement; the policy can succeed there by effectively replaying the trajectory, but it cannot adapt the contact timing, body posture, or force application when the terrain changes.
Using references as a source of robust, terrain-conditioned skill learning, rather than as trajectories to memorize, is therefore itself a central difficulty.
3) Finally, reference-learned skills are usually short-horizon behaviors.
Even if each individual skill works in isolation, a long-horizon task still requires the policy to decide from its own observations \emph{which} skill to invoke, \emph{when} to invoke it, and when to return to ordinary locomotion.

To address these challenges, we present \texttt{Light-Loco-Parkour}, an end-to-end whole-body locomotion system.
On a humanoid robot it traverses demanding terrain, including stairs, narrow beams, boxes, and stepping stones, and performs agile parkour skills including climb-and-step, speed-vault, step-down, and reverse-vault (Fig.~\ref{fig:teaser}).
At its foundation is a perceptive locomotion backbone, which we extend so that the same policy acquires a variety of whole-body skills rather than legged gaits alone.
To overcome data scarcity, each skill begins from a single seed motion that is iteratively refined into a rich set of terrain-paired references covering different obstacle geometries, dynamically feasible by construction.
To chain the skills together, LightLP learns a dedicated transition stage in which the handoffs emerge purely from RL reward, rather than being solved offline by trajectory optimization or a motion graph.
The result is a single policy that, from onboard depth and a velocity command alone, infers and executes the most appropriate skill for the terrain ahead, without an explicit one-hot skill label.

The contributions of this paper are threefold.
\begin{itemize}
  \item \textbf{Whole-body perceptive-control pipeline.}
    We introduce a whole-body perceptive-control pipeline that extends an RL-trained, velocity-tracking locomotion policy with parkour skills learned from object-interacting motions. Conditioned on onboard perception and a velocity command, the unified policy tracks velocity in open terrain, executes whole-body traversal at obstacles, and resumes locomotion afterward.
  \item \textbf{Terrain-conditioned skill acquisition from sparse seeds.}
    Terrain-conditioned skill acquisition grows a single seed motion into a continuum of dynamically feasible, terrain-paired references, for instance expanding one climb reference from a $45$~cm obstacle to $75$~cm, rather than collecting a large motion corpus. Trained on this continuum, one policy per skill generalizes across obstacle geometry.
  \item \textbf{Autonomous skill transition from reward.}
    Reward-driven transition learning lets the decision of \emph{when} to invoke \emph{which} whole-body skill emerge from perception and task reward alone. A single policy switches between locomotion and skills purely from onboard depth and a velocity command, with no one-hot skill label, hand-coded state machine, or runtime motion generator.
\end{itemize}

\section{Related Work}
\label{sec:related}

\subsection{Locomotion}
\label{sec:related:loco}

Classical locomotion is model-based, reducing the robot to a simplified template such as a linear inverted pendulum constrained by the zero-moment point, or a centroidal/rigid-body model solved by model-predictive control and trajectory optimization~\citep{kajita2003biped,orin2013centroidal,kuindersma2016optimization,dicarlo2018dynamic,posa2014direct}.
These models are precise but presume near-flat, coplanar foot contacts and hand-tuned gaits.

Deep reinforcement learning then recast locomotion as a data-driven problem.
The recipe was proven first on quadrupeds.
Hwangbo et al.~\citep{hwangbo2019learning} trained agile locomotion in simulation with an actuator network that closes the sim-to-real gap and transferred it to hardware, while RMA~\citep{kumar2021rma} added rapid online adaptation to changing dynamics.
The same learning paradigm was then carried to bipeds and humanoids, from robust bipedal jumping~\citep{li2302robust} to real-world humanoid walking.
Notably, Radosavovic et al.~\citep{radosavovic2024real} trained a causal transformer over the history of proprioceptive observations with large-scale model-free RL and deployed it zero-shot.
The resulting controller walks across varied outdoor terrain, stays robust to pushes, and adapts in context, all without a hand-designed model.
More recent work even learns an internal denoising world model in place of direct exteroception~\citep{gu2024advancing}.
These controllers, however, act from proprioception alone and remain blind, discovering terrain only on contact.

\emph{Perceptive} locomotion closes this gap by adding an exteroceptive observation that lets the policy choose its motion before contact, and two implementations dominate.
The first builds an explicit elevation map from onboard LiDAR or depth fused with odometry and plans footholds on it~\citep{miki2022learning,wang2025beamdojo,long2025learning}; it is accurate when the map is, but under mapping noise and drift the robot can step into empty space on sparse footholds such as stepping stones or balance beams.
The second is end-to-end, acting directly on raw depth: it sidesteps the mapping pipeline and its latency, at the price of a harder perception problem.
This end-to-end recipe first delivered agile parkour on quadrupeds~\citep{zhuang2023robot,luo2024pie}, and was then carried to humanoids for perceptive locomotion over challenging terrain~\citep{sun2025dpl}.
The two efforts most related to ours combine whole-body skills with perception~\citep{wu2026perceptive,zhang2026learning}.
PHP~\citep{wu2026perceptive} obtains its whole-body locomotion from reference motion that must be synthesized offline by hand, so the robot appears to be magnetically drawn toward any obstacle ahead.
Moreover, it relies on a one-hot command in its observation, which leaves it unable to accurately track the velocity command.
Zhang et al.~\citep{zhang2026learning} instead pair a diffusion-based motion generator with a tracking policy; because both policies must be run at inference, onboard computation becomes slow and an additional compute unit is required, and the approach still inherits the same scarcity of whole-body data, which leaves its motions stiff.

In contrast to these two systems, LightLP differs on three counts.
First, it learns perceptive locomotion from reward alone, with no reference in the loop, so its motion stays natural and robust.
Second, it selects and executes whole-body skills from onboard depth and a velocity command, with no one-hot index, so it follows the commanded velocity faithfully rather than being drawn onto whatever obstacle lies ahead.
Third, it runs as a single policy onboard, without the second network and dedicated compute unit that a runtime motion generator demands.

\subsection{Whole-Body Skills from Human Motion}
\label{sec:related:wbc}

A humanoid shares almost the same morphology as a human, and human motion is among the easiest behavioral data to collect at scale, from motion capture to everyday Internet video.
This precondition, a near-identical body paired with abundant, cheap demonstrations, has spurred a large body of work that leverages human motion data to endow humanoids with whole-body skills.
These methods fall into two categories, separable at the observation level by whether the policy is given an explicit reference.
\emph{Mimic-based} methods~\citep{peng2018deepmimic,liao2025beyondmimic} feed a per-frame reference into the observation and reward the policy for tracking it.
\emph{Style-based} methods~\citep{peng2021amp,mu2025smp} omit the explicit reference, instead matching the distribution of human motion through an adversarial style reward or encoding the reference into a latent token that the policy conditions on.

Peng et al.~\citep{peng2018deepmimic} train a physics-simulated character to track reference motion clips with reinforcement learning under a joint imitation-and-task objective; their reference state initialization (RSI), which resets each episode to a random reference frame, is what makes highly dynamic skills learnable.
Recent systems carry this tracking recipe to real humanoids by closing the sim-to-real gap from different angles.
BeyondMimic~\citep{liao2025beyondmimic} models the actuators from first principles and deploys through a low-latency stack.
ASAP~\citep{he2025asap} learns a residual delta-action model from real-world rollouts to align simulation with the measured dynamics.
KungfuBot~\citep{xie2025kungfubot} pairs physically constrained motion processing with an adaptive tracking curriculum.
Together, these systems transfer agile skills such as cartwheels, kicks, and kung-fu zero-shot to a Unitree G1.
Scaling this idea up, SONIC~\citep{luo2025sonic} trains a single tracking policy on hundreds of hours of motion-capture data with adaptive sampling, yielding a \emph{general} controller that tracks a wide range of motions and generalizes to unseen ones.

The complementary, style-based approach instead matches the \emph{distribution} of reference motion rather than any specific frame: AMP~\citep{peng2021amp} does so through a learned discriminator, and SMP~\citep{mu2025smp} casts the prior as a reusable, task-agnostic score-matching module.
These methods are demonstrated mostly on flat ground, however, and pushing them to acquire whole-body skills tends to deform the motion into unnatural shapes~\citep{wang2025hil}.

What unites both categories is a shared blind spot: these skills make almost no contact with objects off the ground, fundamentally because human references rarely capture a motion together with the scene that shaped it.
The community has therefore pushed to obtain more diverse references.
One intuitive route recovers SMPL motion from video, e.g.\ via GVHMR~\citep{shen2024world}; without known camera intrinsics, however, the recovered motion is systematically biased and produces artifacts such as scene penetration.
NMR~\citep{zhao2026nmr} observes that optimization-based retargeting is non-convex and prone to such artifacts, and instead casts retargeting as learning a data distribution.
It uses clustered reinforcement-learning experts to project noisy human demonstrations onto the robot's feasible manifold, thereby eliminating joint jumps and self-penetration; occlusion, however, remains difficult for video.
Motion capture offers another source: OmniRetarget~\citep{yang2025omniretarget} jointly optimizes the motion together with the manipulated objects and terrain, preserving interactions and generalizing across configurations, but because it optimizes purely at the reference level it cannot guarantee the dynamic feasibility of the resulting motion.

Once references are available, the remaining question is how to learn a skill that generalizes across obstacle geometry rather than overfitting a single demonstration, and the answer turns on the observation.
Deep Whole-body Parkour~\citep{zhuang2026deep} places the reference in the observation and appends a height scan, expecting the scan to let the policy absorb how the obstacle deviates from the reference; in practice this generalizes poorly and its real-robot performance is limited.
HIL~\citep{wang2025hil} instead keeps the reference out of the observation, encoding the terrain through a PointNet latent while still training against a reference-tracking reward; lacking any reference input, the policy converges only with difficulty, and each parkour skill remains tied to one specific terrain, demonstrated without generalization or real-hardware deployment.

Our approach tackles both stages.
For data, we augment a single seed motion, obtained from video or motion capture, into references that are dynamically feasible by construction and exactly matched to the terrain that produced them, rather than kinematic-only or scene-free.
For skill learning, we keep the explicit reference out of the deployable observation, as in HIL, so that each skill generalizes across obstacle geometry; but to overcome the slow, unstable convergence this otherwise incurs, we adopt a teacher--student architecture in which a per-skill expert supervises the student, greatly accelerating convergence.

\section{System Overview}
\label{sec:overview}

\begin{figure*}[!t]
  \centering
  \includegraphics[width=\textwidth]{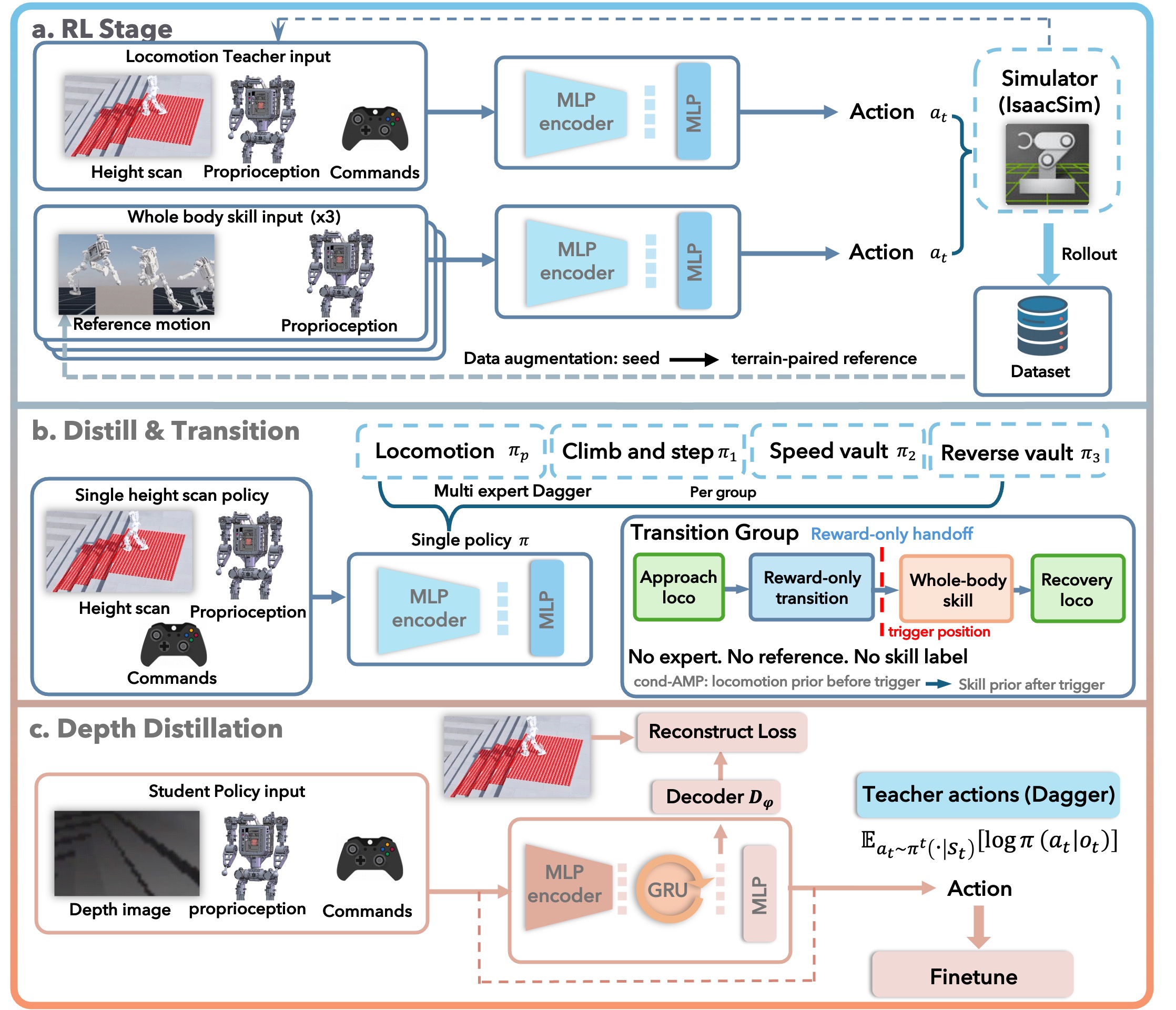}
  \caption{%
    \textbf{Overview of \texttt{Light-Loco-Parkour}.}
    \textbf{(a) RL stage:} using privileged information, we train one perceptive-locomotion teacher and one teacher for each whole-body skill, while data augmentation expands each seed motion into terrain-paired references.
    \textbf{(b) Distill \& transition:} multi-expert DAgger distills the locomotion and skill teachers into a single height-scan policy; a reward-only transition group then fine-tunes the policy to hand off between locomotion and whole-body skills without an expert, reference, or skill label, while the AMP prior switches from locomotion to skill at the trigger position.
    \textbf{(c) Depth distillation:} the resulting height-scan policy is distilled into a recurrent depth policy using onboard-style depth, proprioception, and velocity commands, with teacher-action supervision, auxiliary height-scan reconstruction, and a final fine-tune.}
  \label{fig:pipeline}
\end{figure*}

The overview of LightLP is shown in Fig.~\ref{fig:pipeline} and comprises four modules: perceptive locomotion learning (Section~\ref{sec:method:loco}) and three that together build the whole-body skills, namely data augmentation (Section~\ref{sec:method:data}), skill learning and generalization (Section~\ref{sec:method:distill}), and transition learning (Section~\ref{sec:method:transition}).
A final distillation pass then ports the resulting height-scan policies onto the robot's onboard depth camera for deployment (Section~\ref{sec:method:depth}).

\subsection{Problem Formulation}
\label{sec:overview:formulation}

We train the whole-body perceptive locomotion policy with reinforcement learning, formulating the problem as a partially observable Markov decision process.
At each control step the policy $\pi_\theta(a_t \mid o_t)$ maps an observation $o_t$ to an action $a_t$, and is optimized with PPO~\citep{schulman2017proximal} to maximize the expected discounted return $\mathbb{E}\!\left[\sum_t \gamma^t r_t\right]$ under a reward $r_t$ specified per module.
All training is carried out in simulation in IsaacLab~\citep{mittal2023orbit}.

\subsection{Observation and Action Spaces}
\label{sec:overview:io}

During teacher training, the observation includes privileged information, the velocity command $(v_x, v_y, \omega_z)$, and a local height scan.
The deployable student is then distilled from this height-scan teacher and uses onboard depth in place of the privileged scan.
The action $a_t$ sets target joint positions tracked by a PD controller, and the policy runs onboard at $50$~Hz.

\section{Perceptive Locomotion Learning}
\label{sec:method:loco}

\subsection{Policy Architecture}
\label{sec:method:loco:arch}

The general policy architecture uses an MLP encoder as the core feature extractor for the map input.
A proprioception encoder embeds the proprioceptive observations, and the map encoder embeds the map observations conditioned on this proprioception embedding.
We then feed both the proprioception embedding and the map embedding through a multilayer perceptron (MLP) to output the action.

We instantiate this architecture within a teacher--student scheme~\citep{lee2020learning,miki2022learning,chen2020learning}: a teacher trained on privileged sensing supervises a student that is deployable from onboard observation alone.
The teacher reads a clean height scan.
For its map encoder, we compared a convolutional design with a fully-connected design.
The convolutional encoder yielded no measurable gain yet noticeably lengthened training, so we use an MLP encoder in all experiments.
Since the teacher's network is entirely MLP-based with no recurrence, a single observation carries no memory of the recent past.
We therefore stack the last five observation frames to supply temporal context, following input-history designs that let a policy implicitly infer the latent dynamics and contact state it cannot sense directly~\citep{li2024reinforcement}.
The student, by contrast, is built for onboard deployment, where the privileged quantities available in simulation, most notably the base velocity, can no longer be measured, and where the only exteroception is a single depth image that is noisier and far narrower in field of view than the height scan.
We therefore encode the depth with an MLP, fuse it with the proprioception embedding, pass the result through an RNN, and decode its hidden state with a final MLP into the action.
The recurrence builds memory across frames so that the policy can implicitly infer the unmeasured state, such as the current velocity $(v_x^c, v_y^c, \omega_z^c)$, and retain terrain that has scrolled out of view.

\subsection{Asymmetric Actor and Critic}
\label{sec:method:loco:ac}

We train the teacher with an asymmetric actor--critic~\citep{pinto2018asymmetric,andrychowicz2020learning}, in which the critic accesses privileged information that the actor does not, yielding more accurate value estimates that help the teacher learn better.
Granting privileged input to the teacher's \emph{actor} would speed its learning and yield a stronger-performing teacher, yet it would leave the student, which never sees that input, unable to infer the teacher's observation at distillation and thus unable to reproduce its behavior.
This information gap between an over-informed teacher and a partially observed student is well documented~\citep{walsman2022impossibly,nguyen2023leveraging,warrington2021robust}.
We therefore reason backward from the student when designing the teacher's observation, giving the actor only quantities the student can itself infer~\citep{messikommer2025student}.

In our implementation, the actor reads a noise-free height scan together with proprioception subjected to mild domain randomization.
The sparse, narrow terrains we target, such as stepping stones and balance beams, make precise foot placement essential, so we additionally grant the actor a privileged contact flag.
The critic receives clean, noise-free inputs as well, augmented with oracle information such as under-foot height scans.

\subsection{Environments}
\label{sec:method:loco:env}

\subsubsection{Rewards}
\label{sec:method:loco:rewards}

As a locomotion controller, the policy has two essential objectives: to follow the velocity command and to avoid terminating.
Our reward accordingly comprises two groups, one that rewards tracking of the commanded velocity and one that penalizes the behaviors leading to termination, both summarized in Table~\ref{tab:loco-rewards}.

The tracking group comprises four terms: linear-velocity tracking, angular-velocity tracking, an upright-orientation reward, and a velocity-slack bonus.
Both velocity-tracking terms use an exponential kernel on the tracking error,
\begin{equation}
\label{eq:vel-track}
r_{\mathrm{vel}} = \exp\!\left(-\lVert e\rVert^2/\sigma^2\right),
\end{equation}
where $e$ is either the planar-velocity error $(v_x^c, v_y^c) - (v_x, v_y)$ or the yaw-rate error $\omega_z^c - \omega_z$, and $\sigma$ is the kernel width.
The upright term mixes two kernels on the projected-gravity error $g_{xy}$,
\begin{equation}
\label{eq:orient}
r_{\mathrm{ori}} = \exp\!\left(-2\lVert g_{xy}\rVert^2\right) + 0.1\,\exp\!\left(-\lVert g_{xy}\rVert\right).
\end{equation}

On the hardest terrains, such as stepping stones and balance beams, a policy trained from scratch tends to halt in front of an obstacle to avoid falling.
Under the exponential tracking reward of Eq.~\eqref{eq:vel-track} alone, this timidity is never overcome and the robot never crosses.
A common remedy adds a position-based or goal-reaching reward that pulls the robot forward~\citep{rudin2025parkour,zhuang2023robot}, but our system carries no odometry, which makes a global position reward inapplicable.
We therefore introduce a velocity-slack bonus that rewards the robot whenever its forward speed lies within a band of the command,
\begin{equation}
\label{eq:slack}
r_{\mathrm{slack}} = \mathbf{1}\!\left[\, v_x^c / v_x \in [0.3,\,1.5] \,\right],
\end{equation}
with $v_x$ the commanded forward velocity and $v_x^c$ the current forward velocity.
Unlike the exact tracking above, this loose band lets the policy modulate its speed while negotiating an obstacle, slowing to climb or vault without being penalized, as long as it keeps advancing roughly at the commanded pace.
For the opposite-direction penalty in Table~\ref{tab:loco-rewards}, $\mathbf{v}^c=(v_x^c,v_y^c)$ denotes the current planar velocity and $\hat{\mathbf{v}}=(v_x,v_y)/\lVert(v_x,v_y)\rVert$ denotes the commanded planar direction.

\begin{table*}[t]
\centering
\caption{Reward terms for perceptive locomotion training.}
\label{tab:loco-rewards}
\begin{tabular*}{\textwidth}{@{\extracolsep{\fill}} l c c r}
\hline
\textbf{Term} & \textbf{Expression} & \textbf{Weight} & \textbf{Description} \\
\hline
\multicolumn{4}{l}{\emph{Tracking rewards}}\\
Linear-velocity tracking  & Eq.~\eqref{eq:vel-track} & $+2.0$  & Follow the commanded planar velocity.\\
Angular-velocity tracking & Eq.~\eqref{eq:vel-track} & $+2.0$  & Follow the commanded yaw rate.\\
Upright orientation       & Eq.~\eqref{eq:orient}    & $+1.0$  & Keep the torso upright.\\
Velocity-slack bonus      & Eq.~\eqref{eq:slack} & $+1.5$  & Keep the forward speed near the command.\\
\hline
\multicolumn{4}{l}{\emph{Penalties}}\\
Undesired contact   & $\textstyle\sum_i \mathbf{1}[f_i > 1\,\mathrm{N}]$ & $-2.0$  & Avoid contact on links other than the feet.\\
Joint limit         & $\textstyle\sum_j \mathbf{1}[\,\text{near\,\&\,toward}\,]$ & $-10.0$ & Stay away from joint limits.\\
Illegal footstep    & Eq.~\eqref{eq:illegal}                   & $-1.0$  & Avoid stepping onto a hole or edge.\\
Heading error       & $|\Delta\psi|$                           & $-1.0$  & Follow the commanded heading.\\
Opposite direction  & $\max(0,-\mathbf{v}^c\!\cdot\!\hat{\mathbf{v}})$             & $-1.0$  & Avoid moving against the command.\\
Action rate         & $\lVert a_t - a_{t-1}\rVert^2$           & $-0.1$  & Encourage smooth actions.\\
Foot acceleration   & Eq.~\eqref{eq:foot-acc}                  & $-0.01$ & Land softly with anticipation.\\
\hline
\end{tabular*}
\end{table*}

\begin{figure}[t]
  \centering
  \includegraphics[width=\columnwidth]{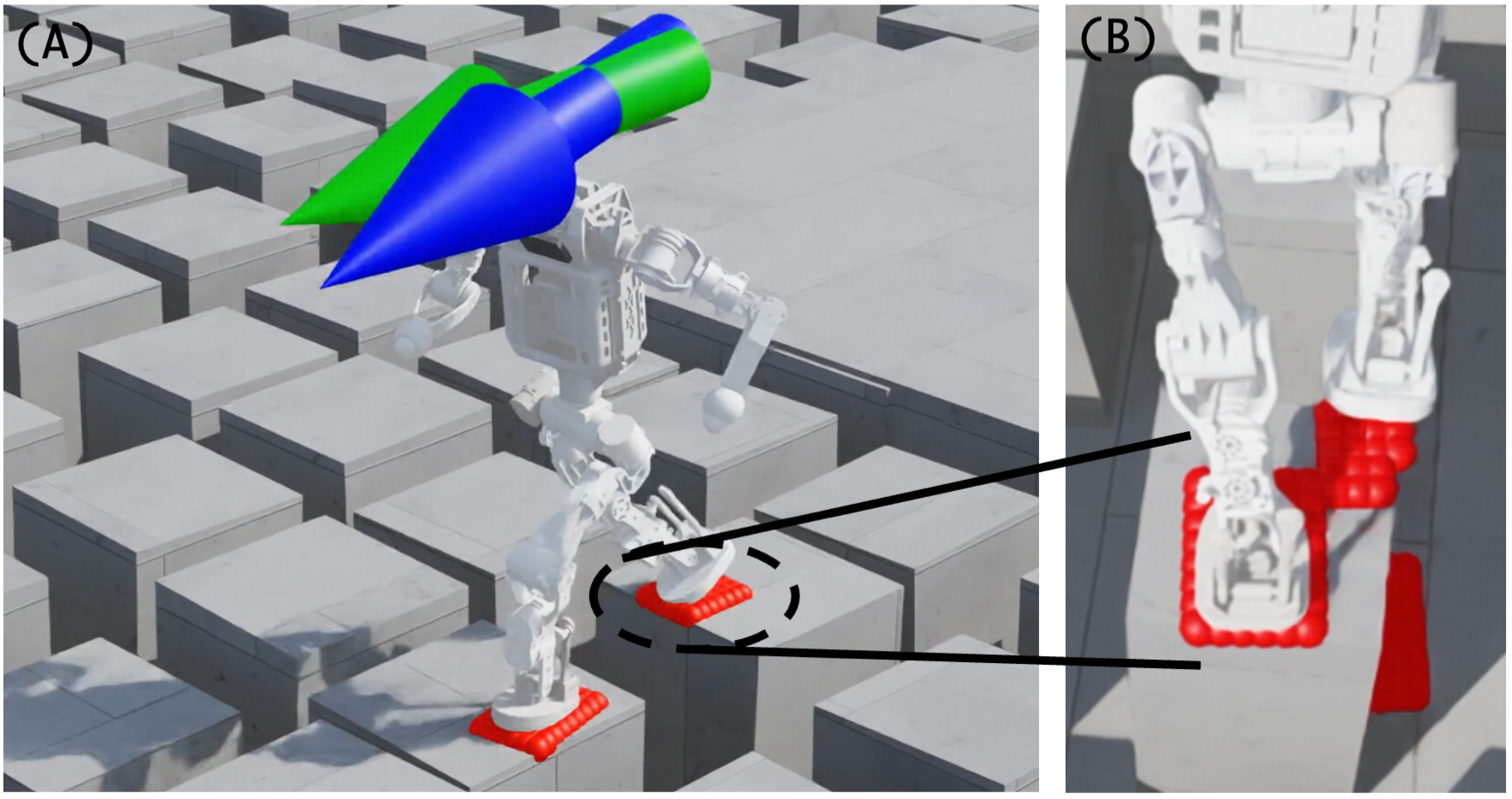}
  \caption{%
    Foot-mounted height scan to compute the illegal-footstep penalty.
    Each foot casts a small grid of downward rays, and a ray whose hit point lies more than $\delta$ below the foot signals that the foot overhangs a hole or an edge, as on the stepping stones shown.}
  \label{fig:stepping-stone}
\end{figure}

With command tracking in place, the second group of terms shapes how well the robot negotiates the terrain.
Most are standard regularization terms, including action-rate smoothing, a joint-limit safety penalty, and an undesired-contact penalty, and are listed in Table~\ref{tab:loco-rewards}.
The remaining two penalties, by contrast, are central to our setting.
The illegal-footstep penalty discourages placing a stance foot over a hole or an edge, which is essential on sparse footholds such as stepping stones and on narrow balance beams.
Rather than building an elevation map~\citep{miki2022learning,wang2025beamdojo,long2025learning}, we mount a small downward height scanner on each foot and read its rays directly.
For each foot in contact, the penalty is the fraction of its rays whose hit point lies more than $\delta=0.1$~m below the foot (Fig.~\ref{fig:stepping-stone}),
\begin{equation}
\label{eq:illegal}
r_{\mathrm{foot}} = \sum_{i} c_i\,\frac{1}{K}\sum_{k=1}^{K}\mathbf{1}\!\left[\, z^{\mathrm{foot}}_i - z^{\mathrm{hit}}_{ik} > \delta \,\right],
\end{equation}
where each contacted foot contributes zero at the center of a stone and grows toward one near a hole or edge, with $c_i$ flagging foot $i$ in contact and $K$ rays per foot.

The foot-acceleration penalty, in turn, shapes how the robot lands.
Its raw signal is the excess linear acceleration of the feet above $\bar a = 30$~m/s$^2$, but instead of acting per step it is accumulated through a first-order exponential-decay filter,
\begin{equation}
\label{eq:foot-acc}
\tilde e_t = \alpha\,\tilde e_{t-1} + \sum_{i\in\text{feet}} \max\!\left(\lVert a_i\rVert - \bar a,\, 0\right), \qquad \alpha = e^{-\Delta t/\tau},
\end{equation}
with $\tau = 0.06$~s, and the term penalizes $\tilde e_t$.
The decaying memory makes a hard impact felt over a short window rather than a single instant, which drives the policy to decelerate each foot before contact and to plan where and how firmly to land.
This turns foothold selection into an anticipatory, perceptive behavior and yields softer, more deliberate steps.

\subsubsection{Termination}
We use the following termination terms, each with a distinct role.
\begin{itemize}
\item \emph{Time-out.} The episode ends at the fixed horizon, marked as a time-out so the value function bootstraps from the final state rather than receiving a failure return.
\item \emph{Out of bounds.} The episode ends when the robot leaves its terrain patch, declared with a $2$~m buffer before the true edge. It bootstraps rather than penalizes, signaling the end of usable terrain rather than a failure.
\item \emph{Joint-velocity guard.} A joint speed beyond $50$~rad/s ends the episode, also treated as a time-out and as a numerical safeguard rather than a behavioral failure.
\item \emph{Torso contact.} A contact above $1$~N on the torso resets the episode, penalizing the collisions and falls we want to avoid.
\item \emph{Excessive acceleration.} A base linear acceleration above $40$~m/s$^2$, measured after a $1$~s warm-up at reset, likewise resets the episode, catching violent impacts that the contact sensor alone may miss.
\item \emph{Fall-over.} When the base tilts past $63^\circ$ from upright the episode terminates, but stochastically at probability $0.01$ per step rather than instantly, leaving an expected hundred-step window in which the policy can attempt to recover its balance.
\end{itemize}

To improve robustness and gather more diverse experience, the two impact triggers, torso contact and excessive acceleration, share a degree of immunity: a random $10\%$ of the environments, resampled every $200$ steps, ignore both and continue through brief contacts or impacts rather than reset on the first frame. We expose this immunity flag to the critic (Section~\ref{sec:method:loco:ac}), so value estimation stays consistent with whether an environment can actually terminate on impact.

\subsubsection{Terrain and Curriculum}
\begin{figure*}[!t]
  \centering
  \includegraphics[width=\textwidth]{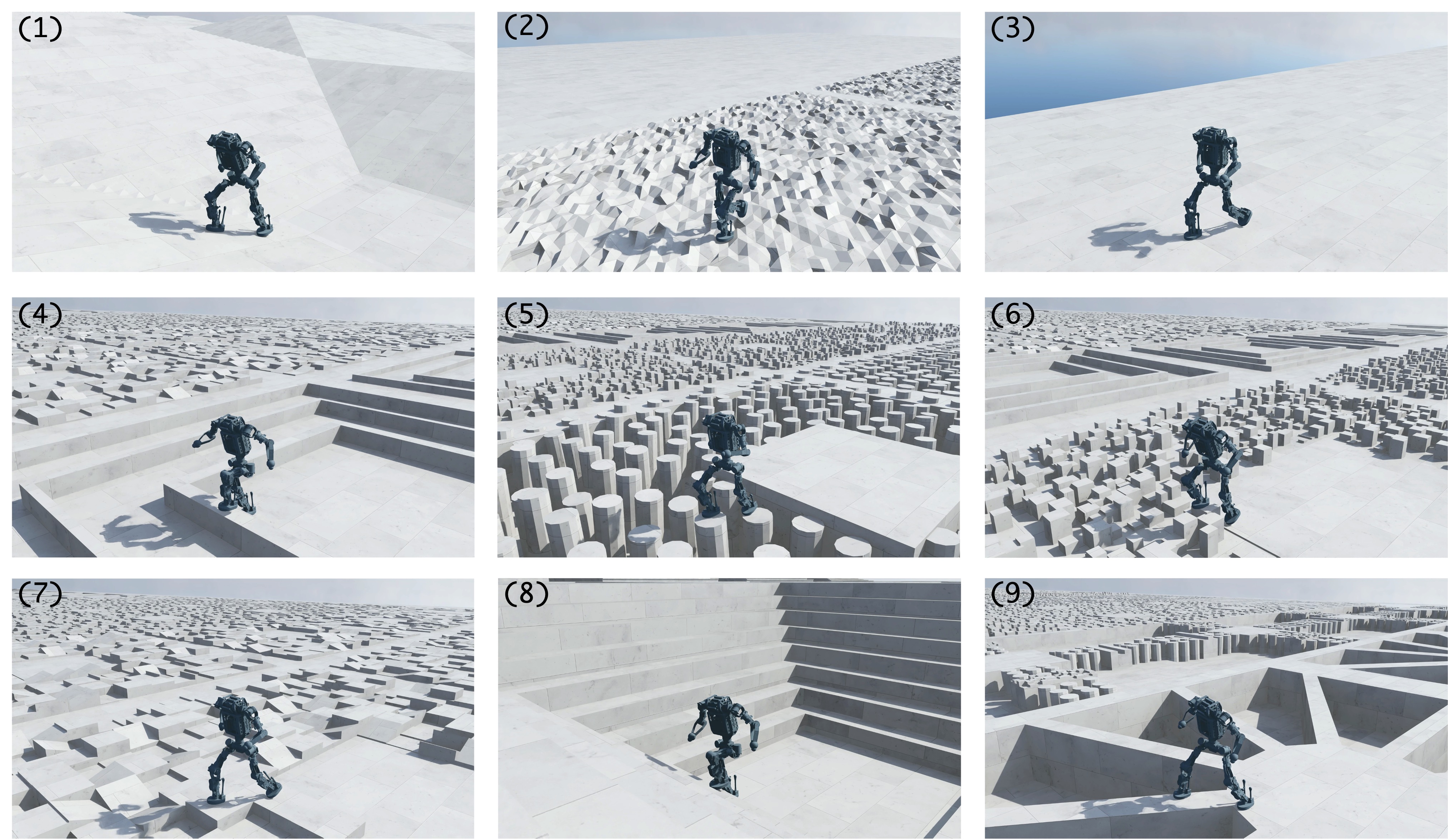}
  \caption{%
    The locomotion training terrain types. Each column is a distinct terrain type and difficulty increases along the rows:
    (1) slopes,
    (2) Perlin-noise rough ground,
    (3) flat ground,
    (4) rails,
    (5) raised pillars,
    (6) stepping stones,
    (7) box fields,
    (8) stairs, and
    (9) plank bridges.}
  \label{fig:terrain}
\end{figure*}

We train on a single procedurally generated terrain that spans a broad range of obstacles, arranged as a grid of $10$ difficulty levels and $32$ columns of terrain type (Fig.~\ref{fig:terrain}).

Among them, the sparse-foothold terrains, including stepping stones, plank bridges, and pillar fields, make precise foot placement essential, and they are where the illegal-footstep penalty and the foot-mounted height scan of Section~\ref{sec:method:loco:rewards} are especially important.

Difficulty rises along the rows, and we move each robot through the levels with a terrain-level curriculum~\citep{rudin2022learning}.
A robot that, under a velocity command, walks more than half a terrain cell while tracking that command well is promoted to a harder level. One whose tracking falls short is demoted to an easier one.
Distance here is the episode-cumulative path length rather than straight-line displacement, so a command that resamples mid-episode does not corrupt the signal.
A further $10\%$ of resets are placed at a random level regardless of performance, which keeps every difficulty populated and prevents the early levels from emptying as the policy improves.

\subsubsection{Events and Domain Randomization}
\label{sec:method:loco:dr}
The policy is trained entirely in simulation, so we randomize the quantities the simulator cannot match to the real robot and add disturbances the robot will meet in the field~\citep{peng2018sim}.
Table~\ref{tab:dr} lists the ranges, grouped by what they model.
Ground-contact terms cover the friction and restitution of every body. Inertial terms perturb link masses, add an unmodeled torso payload, and shift the base center of mass. Actuation terms vary armature, joint friction, and the coupled stiffness--damping that stands in for motor-constant ($K_t$) variation. These last terms are resampled slowly, no more than once every $5\!\times\!10^4$ steps, so that each episode sees a temporally consistent actuator rather than gains that jump within a rollout.

We highlight two terms.
We perturb the depth camera's mounting pose at every reset~\citep{tobin2017domain}, since the student policy reads this camera at deployment and a fixed extrinsic calibration is never exactly right on hardware.
We also push the base with a random velocity every few seconds, which forces the policy to recover its balance from disturbances rather than relying on an undisturbed gait.
Finally, each episode begins from a randomized base pose and velocity and from perturbed joint positions, so the policy is exposed to a wide range of starting states rather than a single nominal stance.

\begin{table}[t]
\caption{Domain randomization ranges.}
\label{tab:dr}
\centering
\renewcommand{\arraystretch}{1.2}
\begin{tabular}{@{}l l@{}}
\hline
Parameter & Range \\
\hline
\multicolumn{2}{@{}l}{\emph{Ground contact} (startup)} \\
\quad Friction (static, dynamic)   & $\times\,[0.2,\,1.3]$ \\
\quad Restitution                  & $[0.0,\,0.8]$ \\
\multicolumn{2}{@{}l}{\emph{Inertial properties} (startup)} \\
\quad Link mass                    & $\times\,[0.85,\,1.15]$ \\
\quad Torso payload                & $\times\,[1.0,\,1.2]$ \\
\quad Base center of mass          & $\pm 2.5$~cm (x, y, z) \\
\multicolumn{2}{@{}l}{\emph{Actuation} (slow resample)} \\
\quad Joint armature               & $\times\,[0.75,\,1.25]$ \\
\quad Coulomb friction             & $\times\,[0.7,\,1.3]$ \\
\quad Viscous friction             & $+\,[0.0,\,0.05]$ \\
\quad Stiffness--damping ($K_t$)   & $\times\,[0.875,\,1.075]$ \\
\multicolumn{2}{@{}l}{\emph{Perception} (reset)} \\
\quad Depth-camera position        & $\pm 1$~cm \\
\quad Depth-camera orientation     & $\pm 0.025$~rad \\
\multicolumn{2}{@{}l}{\emph{Disturbance} (interval)} \\
\quad Push interval                & $3.7$--$4.2$~s \\
\quad Push velocity                & $\pm 0.5$~m/s \\
\multicolumn{2}{@{}l}{\emph{Initialization} (reset)} \\
\quad Base position, yaw           & $\pm 0.5$~m,\ $\pm\pi$ \\
\quad Base linear, angular vel.    & $\pm 0.5$ \\
\quad Joint position, velocity     & $\pm 0.1$~rad,\ $\pm 1.0$~rad/s \\
\hline
\end{tabular}
\end{table}

\section{Whole-Body Skills}
\label{sec:method:wbs}

We build whole-body parkour skills that recruit not only the legs but also the arms and upper body to perform maneuvers such as climb-and-step, speed-vault, and reverse-vault, expanding the range of terrain the robot can reach.

\subsection{Data Augmentation}
\label{sec:method:data}

To address the scarcity of whole-body references with environmental contact, we build a seed dataset from short Internet videos of a human performing each target skill.
For each video, we extract the per-frame SMPL body model with GVHMR~\citep{shen2024world} and retarget it with GMR~\citep{joao2025gmr}.
We then manually place a virtual obstacle at the contact points implied by the retargeted hands and feet, yielding a coarse motion-and-terrain seed pair.
However, neither GVHMR nor GMR is aware of the obstacle, so the seed reference inevitably collides with it: the supporting hand sinks into the top face, and a knee clips through the front edge.
The simulator, by contrast, is itself an effective retargeting resource: a reference tracked under physics comes out both in seamless contact with the obstacle and dynamically feasible~\citep{xu2025parc,kim2026flip}.

\subsubsection{Object-Interaction Mimic}
\label{sec:method:oim}
\begin{figure*}[t]
  \centering
  \includegraphics[width=\textwidth]{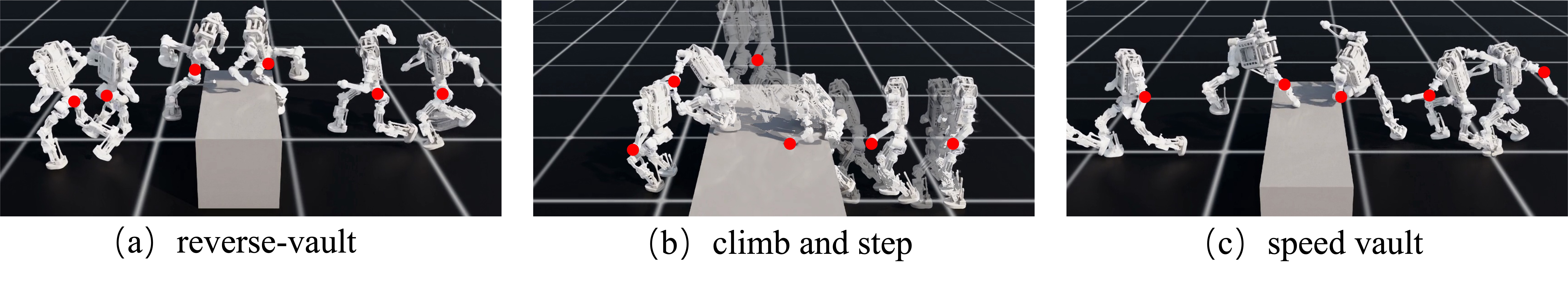}
  \caption{%
    Red markers indicate the body where the global body position reward is applied: the wrist for climb-and-step, the elbow for speed-vault, and the pelvis for reverse-vault.}
  \label{fig:penetration}
\end{figure*}

Our mimic stage builds on BeyondMimic~\citep{liao2025beyondmimic}, which we extend from free-space imitation to contact-rich, object-interacting skills.
Since a parkour skill is defined by where the body meets the obstacle rather than by the pose alone, we track the global root position instead of only the root-relative pose, anchoring every contact, whether from the hand, foot, or pelvis, to a fixed location in the scene.
We also add two privileged observations, the distance to the obstacle and its size, so the policy can tell how far the true surface lies from an imperfect seed and correct toward it even when that seed still penetrates the obstacle.
We then adapt the reward: for the body part meant to make contact, a global-frame position tracking term insists the contact point actually reach the obstacle surface~\citep{xie2025kungfubot}, turning a nominal contact into an enforced physical one.
Reference State Initialization~\citep{peng2018deepmimic}, which resets each episode to a random reference frame, is essential here: it spreads exploration across the whole motion and exposes the policy to the high-momentum mid-skill states, such as the apex of a vault and the push-off of a climb, without which these dynamic skills fail to emerge.
In the end, this yields a reference that matches the obstacle exactly (Fig.~\ref{fig:penetration}).

\subsubsection{Iterative Self-Augmentation}
Once the policy reliably traverses the seed obstacle, we roll out its trajectory in the simulator and adopt it as the next-iteration reference.
Because the rollout is produced by the policy under the physics engine and the robot's actuator limits, it is dynamically and kinematically feasible by construction, even though its kinematic shape gradually drifts from the original SMPL trajectory across iterations.
At each iteration, we lift both the obstacle and the motion by 5--10~cm so that the next round extracts a slightly harder version of the same skill.
Formally, at iteration $i$ the loop produces the next motion-and-terrain pair via
\begin{align}
\label{eq:refinement}
\pi_i &= \arg\min_{\pi}\, \mathcal{L}_{\text{mimic}}(\pi;\, r_i, d_i), \\
\hat{r}_i^{(k)} &\sim \mathrm{Rollout}(\pi_i,\, d_i), \quad k \in [K], \nonumber \\
\hat{r}_i^{\star} &= \arg\max_{k \in [K]}\, \mathrm{Score}\big(\hat{r}_i^{(k)},\, r_i\big), \nonumber \\
(r_{i+1},\, d_{i+1}) &= \big(\hat{r}_i^{\star} + m\,\mathbf{e}_z,\;\; d_i + m\,\mathbf{e}_z\big), \nonumber
\end{align}
where $\mathbf{e}_z$ is the gravity-axis unit vector, $m \in [5, 10]$~cm is the per-iteration lift, and $\mathcal{L}_{\text{mimic}}$ is a DeepMimic-style tracking loss optimized by the \textsc{MimicTrain} step of Algorithm~\ref{alg:refinement}.
We store each rollout reference together with its paired terrain, and the accumulated pairs form the augmented dataset used for skill learning.

\begin{algorithm}[t]
\caption{Iterative Reference Refinement}
\label{alg:refinement}
\begin{algorithmic}[1]
\State \textbf{Definition:} $r_i$: reference motion at iteration $i$; $d_i$: paired obstacle; $\pi_i$: policy trained at iteration $i$; $m$: per-iteration height increment; $K$: number of parallel rollouts; $N$: number of iterations.
\State \textbf{Input:} seed pair $(r_0, d_0)$ recovered from an Internet video via GVHMR, with $d_0$ manually aligned to the human contacts in $r_0$.
\State $\mathcal{R} \leftarrow \{(r_0, d_0)\}$
\For{$i = 0, 1, \dots, N-1$}
    \State $\pi_i \leftarrow \textsc{MimicTrain}(r_i, d_i)$
    \State $\{\hat{r}_i^{(1)}, \dots, \hat{r}_i^{(K)}\} \leftarrow \textsc{Rollout}(\pi_i, d_i, K)$
    \State $\hat{r}_i^{\star} \leftarrow \arg\max_{k}\ \textsc{Score}(\hat{r}_i^{(k)}, r_i)$
    \State $r_{i+1} \leftarrow \textsc{Lift}(\hat{r}_i^{\star}, m)$, \quad $d_{i+1} \leftarrow \textsc{Lift}(d_i, m)$
    \State $\mathcal{R} \leftarrow \mathcal{R} \cup \{(r_{i+1}, d_{i+1})\}$
\EndFor
\State \textbf{Return:} $\mathcal{R} = \{(r_0, d_0), \dots, (r_N, d_N)\}$
\end{algorithmic}
\end{algorithm}

\subsection{Skill Learning and Generalization}
\label{sec:method:distill}

\subsubsection{Height-Scan Distillation}
Although the Object-Interaction Mimic stage (Section~\ref{sec:method:oim}) yields a feasible reference for each skill, the resulting experts are inherently neither general nor deployable.
The mimic expert is fed the obstacle geometry directly, and, lacking odometry, it has no way to anchor itself to a global reference frame.
Each expert therefore reproduces a single motion tied to one obstacle, rather than a behavior that transfers across geometry or runs from onboard sensing alone.
What we want instead is a whole-body skill that runs on the same observation as our perceptive locomotion policy: proprioception, a local height scan, and the velocity command.
Exteroception and the velocity command then serve as a conditional input that tells the policy which task to perform at each moment, without any explicit skill label.

From the continuum of references produced by data augmentation we pick, for each skill, the iteration whose obstacle is the most demanding: furthest, tallest, and closest to the joint torque and angular-velocity limits of the deployment hardware.
We distill this hardest expert directly onto the perceptive locomotion observation.
The hardest expert is the most demanding case for distillation from exteroception alone, so using it lets the distilled policy inherit the full operating envelope rather than only the easy interior of the curriculum.

We distill each chosen expert with DAgger~\citep{ross2011reduction},
collecting on-policy rollouts from the student and supervising them with the expert's actions at the matching states.
Distilling on the student's own sampled states alone, however, yields a brittle policy: once a rollout drifts off the expert's distribution, imitation offers no signal for recovering, an instability also reported in HIL~\citep{wang2025hil}.
We therefore open a PPO loss alongside the imitation term, so the policy retains the ability to recover even when its tracking is imperfect.
The final objective thus combines expert distillation with PPO,
\begin{align}
\label{eq:distill-loss}
\mathcal{L}_{\text{total}} &= \mathcal{L}_{\text{DAgger}} + \lambda_{\text{RL}}\,\mathcal{L}_{\text{PPO}}(r), \\
r &= w_{\text{task}}\,r_{\text{task}} + w_{\text{goal}}\,r_{\text{goal}} + w_{\text{mimic}}\,r_{\text{mimic}}, \nonumber
\end{align}
where the imitation term
\begin{equation}
\label{eq:dagger-loss}
\mathcal{L}_{\text{DAgger}}(\pi_\theta) \;=\; \mathbb{E}_{s \sim d^{\pi_\theta}}\!\left[\,\left\| \pi_E(s_{\text{priv}}) - \pi_\theta(s^{\text{stu}}) \right\|_2^{\,2}\,\right]
\end{equation}
penalizes the gap between the expert action under the privileged observation $s_{\text{priv}}$ and the student action under the deployable observation $s^{\text{stu}}$, evaluated on the student-visited states $d^{\pi_\theta}$, $\mathcal{L}_{\text{PPO}}$ is the standard clipped surrogate driven by the composite reward $r$, and $\lambda_{\text{RL}}$ trades off the two.

The reward, domain randomization, and terminations are inherited from the Object-Interaction Mimic stage (Section~\ref{sec:method:oim}), and only the observation changes to the height scan and velocity command.
Table~\ref{tab:distill-spec} gives the three terms: a goal reward on the terminal base pose $(p_T, \theta_T)$ relative to its target $(p^{\star}, \theta^{\star})$, a mimic reward on the joint angles $q_t$ against the reference $\hat q_t$, and an early termination once the base $p_t$ drifts from the reference $\hat p_t$.

\begin{table}[t]
\caption{Reward and termination terms inherited from the Object-Interaction Mimic stage (Section~\ref{sec:method:oim}).}
\label{tab:distill-spec}
\centering
\renewcommand{\arraystretch}{1.3}
\begin{tabular}{@{}l l@{}}
\hline
Term & Expression \\
\hline
Goal reward $r_{\text{goal}}$            & $-\lVert p_T - p^{\star}\rVert_2 - \alpha\,\lVert \theta_T - \theta^{\star}\rVert$ \\
Mimic reward $r_{\text{mimic}}$          & $\exp\!\big(-\beta\,\lVert q_t - \hat q_t\rVert_2^{\,2}\big)$ \\
Reference-deviation termination          & $\lVert p_t - \hat p_t\rVert_2 > 25$~cm \\
\hline
\end{tabular}
\end{table}

\subsubsection{Skill Generalization}
After distillation, the policy $\pi_i$ reproduces its skill only on the single obstacle its expert was trained on.
The goal of this stage is to turn that single-configuration policy into one that handles the whole range of obstacles collected during data augmentation.
We continue training $\pi_i$ across that range, but now without an expert to imitate.
Dropping the expert, and asking for generalization rather than reproduction, drives the changes below.

\emph{Terrain.}
Each skill trains on its own terrain, whose single obstacle is paired one-to-one with the corresponding motion reference recovered during data augmentation.
We randomize the obstacle's placement in the ground plane and arrange the configurations into a curriculum of difficulty levels, so the policy sees the full range of placements rather than the single one its expert was trained on.

\emph{Reward.}
The mimic reward now tracks the motion paired with the currently sampled obstacle, so as the curriculum varies the terrain the imitation target follows the matching reference rather than a single fixed motion.
On its own, though, kinematic imitation no longer suffices once the expert supervision is gone: the policy can satisfy the mimic term yet still drift away from actually crossing the obstacle.
We therefore add a task reward that pulls the robot toward a target $p^{\star}$ placed a fixed offset beyond the obstacle,
\begin{equation}
\label{eq:task-gen}
r_{\text{task}} = \exp\!\big(-\lVert p^{\star} - p_t \rVert_2 / \sigma_c\big),
\end{equation}
where $p_t$ is the robot's anchor position.
The target is defined relative to the obstacle, which the height scan perceives, so this reward needs no odometry.
The continuous proximity term keeps the robot committed to crossing rather than stalling in front of the obstacle.
We pair it with a terminal reward of the same form that switches on only once the reference motion has finished playing,
\begin{equation}
\label{eq:term-gen}
r_{\text{term}} = \exp\!\big(-\lVert p^{\star} - p_t \rVert_2 / \sigma_t\big)\,\mathbf{1}[\text{done}], \qquad \sigma_t < \sigma_c,
\end{equation}
where $\mathbf{1}[\text{done}]$ marks the end of the motion playback rather than a requirement to reach the target.
Its sharper kernel then rewards being close to the target once the motion is over.

\emph{Initialization.}
Since the obstacle now changes from episode to episode, we disable Reference State Initialization and instead reset the robot at the first frame of the motion, just before the obstacle.
We keep that first-frame base pose but draw the joint angles from the perceptive locomotion policy rather than from the reference.
Starting each skill from a locomotion-consistent posture makes the policy more robust and eases the later handoff between locomotion and skills during transition training (Section~\ref{sec:method:transition}).

\subsection{Transition Learning}
\label{sec:method:transition}

At this point we have a perceptive locomotion policy $\pi_p$ (Section~\ref{sec:method:loco}) and a set of whole-body skills $\{\pi_i\}$, each trained in isolation on its own obstacle.
What the system still lacks is any way to choose among them.
Since every skill was trained separately, the locomotion policy cannot invoke them: confronted with an obstacle, it falls back on locomotion alone and tries to walk through, which on the harder terrains it usually cannot.
The skills, for their part, know how to traverse an obstacle but not when to begin or end one, as the references behind them cover only the traversal and not the approach or exit.
Closing this gap is the goal of the transition stage: deciding when to call a skill and when to keep walking, without a hand-coded state machine.

We close this gap in two steps, following the multi-expert distillation and RL fine-tuning of Parkour in the Wild~\citep{rudin2025parkour}: we first merge the policies into one, then fine-tune that single policy to switch between them.

\subsubsection{Multi-Expert Distillation}
We build one combined environment partitioned into groups and assign a subset of robots to each:
\textbf{(a)} a \emph{locomotion group} carrying the same terrain used to train $\pi_p$ (Section~\ref{sec:method:loco}), and
\textbf{(b)} one \emph{skill group} per skill, carrying that skill's terrain (Section~\ref{sec:method:distill}).
A single student is supervised by the matching expert, either $\pi_p$ or the corresponding $\pi_i$, in each group through DAgger alone, with no PPO term, exactly as in the height-scan distillation above.
The student is never given a skill label: to act correctly it must read the terrain from its exteroception and reproduce the behavior of whichever expert owns that group.
The critic, by contrast, does receive a one-hot indicator of the current group, so that value estimation can account for the differing rewards across groups while the actor stays label-free and deployable.
Unlike quadruped parkour~\citep{rudin2025parkour}, where the experts differ mainly in where the feet land, our skills are whole-body, so the student must also reproduce the arm and torso motion that drives each skill against the obstacle, not merely the foothold pattern.

\subsubsection{Transition Fine-Tuning}
\begin{figure*}[t]
  \centering
  \includegraphics[width=\textwidth]{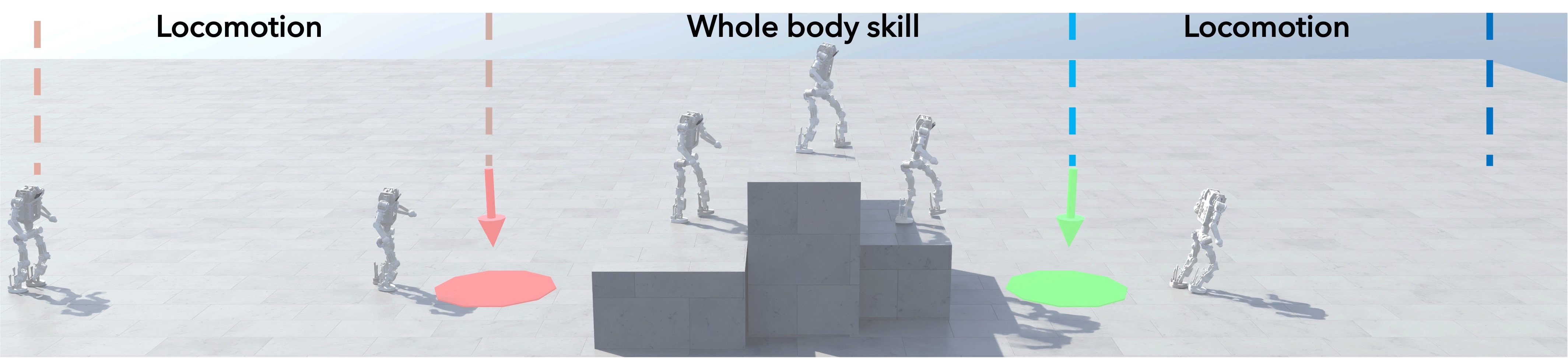}
  \caption{    Transition group. The robot is reset across all three regions, namely the locomotion terrain, the whole-body-skill zone, and the far side of the obstacle, so every phase of a crossing is trained. The conditional motion prior switches from the locomotion AMP to the skill AMP once the robot passes the trigger position.
  }
  \label{fig:transition}
\end{figure*}
Distillation yields a policy that executes each behavior on its own terrain, yet it still cannot \emph{transition}. Nothing in the combined environment ever asks it to leave one behavior and enter another within a single rollout.
We therefore fine-tune the distilled policy with RL, adding a third group to the two above.
\textbf{(c)} \emph{Transition group}: cluttered scenes that interleave the skill obstacles with the open locomotion terrain, with no per-frame reference motion.
A hand-shaped reward that explicitly tells the robot when to vault, climb, or walk fails here because the right choice is highly state-dependent, and any miscalibration makes the policy commit to a skill too early or refuse to engage at all.
We instead drop the mimic reward, since there is no reference to mimic across a transition, and give group~(c) its own reward,
\begin{equation}
\label{eq:transition-reward}
\begin{split}
r_{\text{trans}}
&=
w_v r_{v} + w_{\omega} r_{\omega_z} + w_{\text{pos}} r_{\text{pos}} \\
&\quad + w_{\text{prox}} r_{\text{prox}} + w_{\text{amp}} r_{\text{amp}},
\end{split}
\end{equation}
made of four kinds of term.
A \emph{task} reward tracks the commanded linear velocity $r_v$ and yaw rate $r_{\omega_z}$.
A \emph{sparse position} reward $r_{\text{pos}}$ pays out only once the robot reaches a target placed \emph{behind} the obstacle, marking a completed crossing.
A \emph{dense proximity} reward $r_{\text{prox}}$ grows as the robot closes on that target, supplying the gradient that the sparse term lacks.
Finally, a \emph{conditional} adversarial motion-prior reward~\citep{peng2021amp},
\begin{equation}
\label{eq:amp-reward}
r_{\text{amp}}(s_t, s_{t+1}) \;=\; -\log\!\bigl(1 - D_\phi(s_t, s_{t+1})\bigr),
\end{equation}
keeps the motion close to the learned behaviors, with the discriminator $D_\phi$ switched by phase: before the robot reaches the skill's trigger position the prior is the \emph{locomotion} AMP, and once it passes that position the prior switches to the \emph{skill} AMP.
This phase switch is what actively drives the handoff: the robot is pushed to walk up to the obstacle and then, at the trigger, to adopt the skill that carries it across (Fig.~\ref{fig:transition}).
Unlike Parkour in the Wild, which feeds a goal into the observation and lets transitions emerge as the robot drives toward it~\citep{rudin2025parkour}, our policy observes only a velocity command, so it is this conditional prior, rather than a goal, that elicits the switch.

Because the position reward is sparse and paid only behind the obstacle, a policy that has never reached the far side receives no useful learning signal to guide it there.
We therefore reset robots not only in front of the obstacle but also behind it, so that some episodes begin already past the crossing and immediately collect the terminal position reward, seeding the exploration that the front-reset episodes must otherwise discover on their own (Fig.~\ref{fig:transition}).

Throughout the fine-tune, groups (a) and (b) act as \emph{anchors}: their dense, well-shaped rewards keep the policy from drifting away from either competent locomotion or any of the learned skills, while group (c) is where the transition behavior is acquired under the sparse reward of Eq.~\eqref{eq:transition-reward}.
The policy is never told which skill to call: when the exteroception reports an obstacle that matches a known skill, its behavior gravitates toward the corresponding $\pi_i$, as that is the only way to satisfy both $r_{\text{amp}}$ and $r_{\text{pos}}$.
When the obstacle matches no skill, the policy falls back to $\pi_p$ and either bypasses the obstacle or slows down.
The resulting policy switches between locomotion and skills smoothly, without explicit gating, and recovers gracefully when an attempted skill aborts mid-execution.

\section{From Height Scan to Onboard Depth}
\label{sec:method:depth}

Up to this point every policy operates on a height scan, which is convenient in simulation: it offers wide lateral coverage and is free of self-occlusion.
The corresponding deployment-time observation, however, is a single chest-mounted depth image with a limited field of view, and a substantial fraction of the obstacle frequently leaves the camera frustum mid-skill, for instance when the torso pitches forward over a vault and the camera ends up looking at the sky or at the obstacle's edge.
To bridge this gap, we run a final distillation pass from each height-scan policy onto a recurrent depth-conditioned policy under the same DAgger + PPO objective used to learn the skills.
The recurrent stage feeds depth features into a GRU, so the policy can carry a short-term memory of terrain that has just fallen out of view rather than acting only on the current frame.

To make that memory encode terrain explicitly, we add an auxiliary reconstruction loss: a small decoder maps the GRU hidden state $h_t$ back to the privileged height scan $s^{\text{scan}}_t$ that the teacher observed but the student does not,
\begin{equation}
\label{eq:recon-loss}
\mathcal{L}_{\text{recon}} = \big\lVert D_\psi(h_t) - s^{\text{scan}}_t \big\rVert_2^2,
\end{equation}
where $D_\psi$ is a lightweight reconstruction decoder with a bottleneck latent representation.
Used as an auxiliary training loss, $\mathcal{L}_{\text{recon}}$ forces the recurrent state to accumulate a height-scan-like representation of the surrounding terrain from the stream of partial, self-occluded depth frames. The decoder is used only at training time and is dropped at deployment.

Distilling onto clean simulated depth would transfer poorly, as the hardware RealSense D435 returns far noisier and slower images.
We therefore render the simulated depth through a noise-and-latency model calibrated to the physical camera.
Each frame is corrupted by range-dependent Gaussian noise, with standard deviation $0.005 + 0.02\,d$ for a measured depth $d$ (in meters), and a global depth-scale jitter of $\pm 5\%$.
On top of this we apply persistent block dropout: small patches, refreshed every few frames, are set to the camera's maximum range to mimic the invalid regions the D435 produces, biased toward the top and bottom rows where they concentrate on the real sensor.
We also replay the camera's measured timing, with a $30$--$60$~ms processing latency at a $27$--$33$~Hz frame rate, so the policy learns to act on stale, low-rate depth rather than the instantaneous frames available in simulation.
Finally, a short fine-tune under this noise-and-latency model yields the policy we deploy on hardware.

\section{Experimental Results}
\label{sec:result}

\subsection{Experimental Platform: Lightbot~0}
\label{sec:exp:platform}

\begin{figure}[t]
  \centering
  \includegraphics[width=\columnwidth]{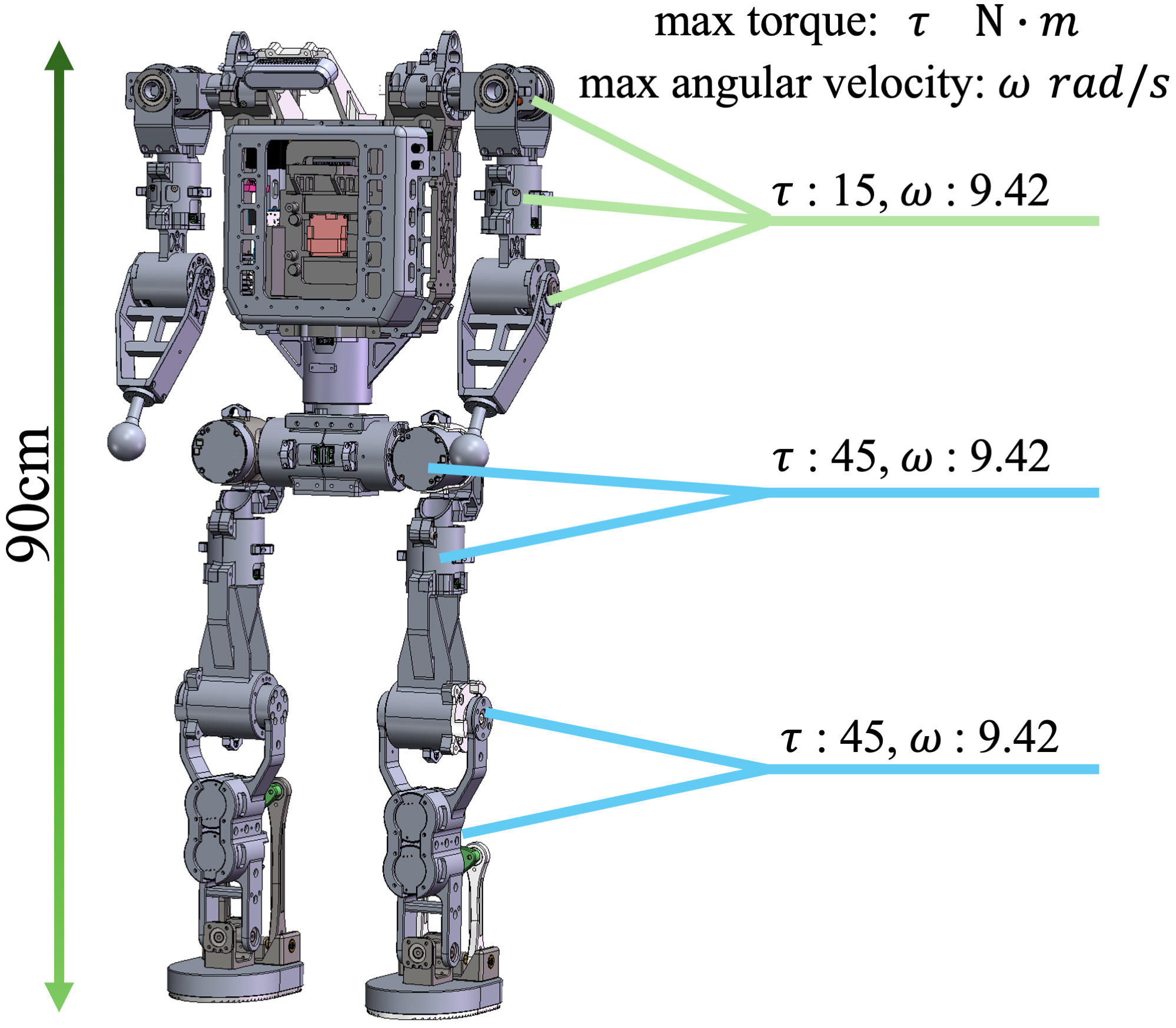}
  \caption{%
    \textbf{Lightbot~0}, the custom-built $90$~cm humanoid used in all our experiments, with $21$ actuated degrees of freedom, a chest-mounted depth camera, and a pelvis IMU.}
  \label{fig:liangzai}
\end{figure}

All experiments are conducted on \textbf{Lightbot~0}, a custom-built compact humanoid developed at Light Origins (Fig.~\ref{fig:liangzai}).

The platform stands $90$~cm tall, weighs $18.9$~kg, and exposes $21$ actuated degrees of freedom distributed across the legs, torso, and arms, with parallel four-bar ankle linkages on both legs.
Joint actuation is supplied by quasi-direct-drive motors with two torque tiers, $45$~N$\cdot$m for the waist and lower-body joints and $15$~N$\cdot$m for the arm joints, and a uniform peak angular velocity of $9.42$~rad/s.
These limits are substantially below those of the $1.3$~m-class platforms used in concurrent humanoid parkour work~\citep{wu2026perceptive,zhang2026learning,zhuang2026deep};
in particular, the limited actuation is the reason the data-and-curriculum pipeline of Section~\ref{sec:method:data} cannot rely on direct mocap retargeting from full-size humans.

The onboard sensor stack consists of a chest-mounted Intel RealSense D435 depth camera, tilted $30^\circ$ downward to keep the terrain just ahead of the feet in view, together with a 6-axis IMU at the pelvis and joint encoders.
All policies reported in this paper run from this onboard stack alone, with no LiDAR, motion capture, or external state estimation.
The deployed policy executes at $50$~Hz on an onboard NVIDIA Jetson Orin Nano, a compact $7$--$25$~W edge module rated at $67$~INT8~TOPS, which is a far lighter compute budget than the dedicated units required by methods that run a motion generator at inference~\citep{zhang2026learning}.

\subsection{Real-World Experiments}
\label{sec:exp:real}
\begin{figure*}[t]
  \centering
  \includegraphics[width=\textwidth]{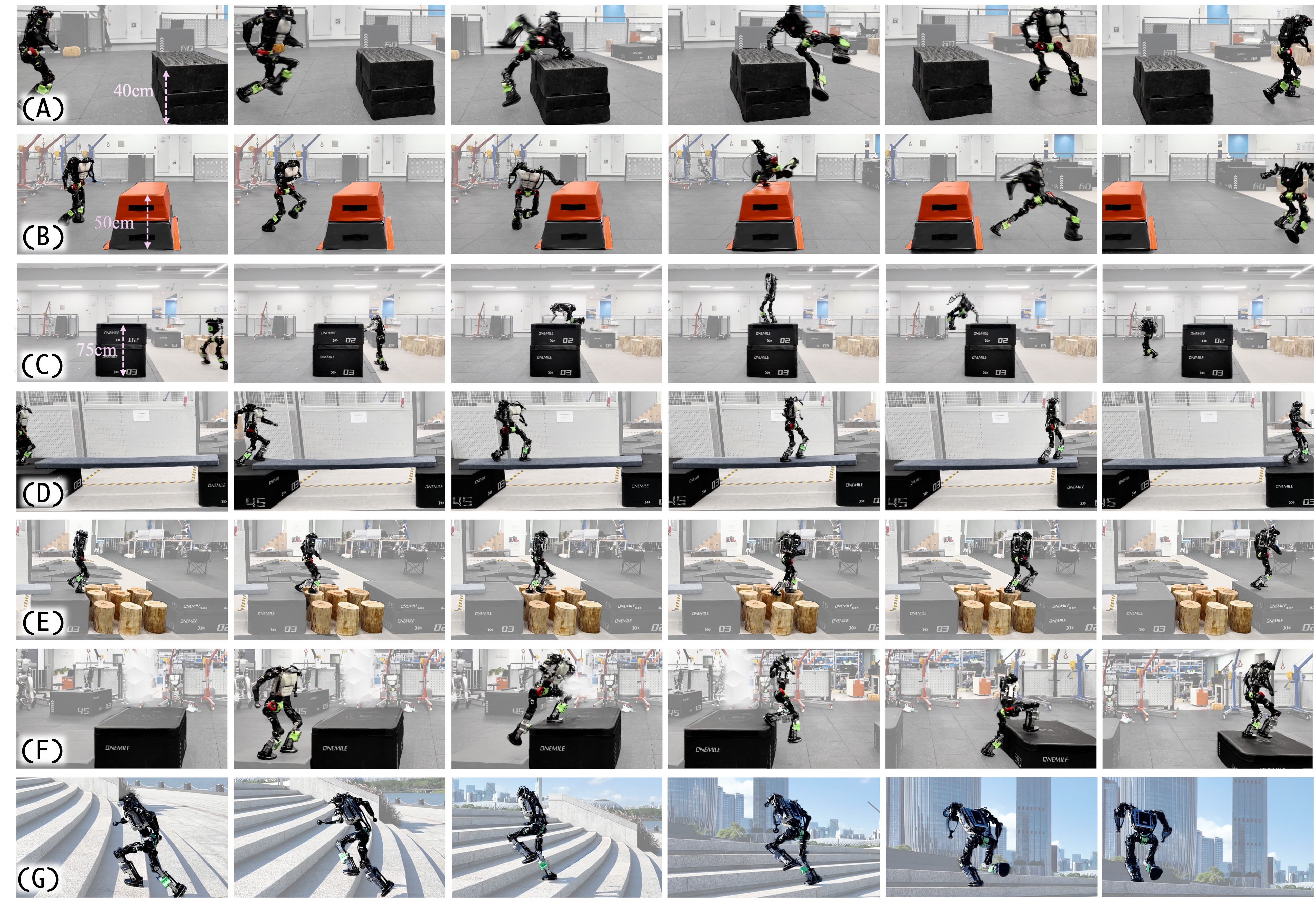}
  \caption{%
    \textbf{Real-world deployment on Lightbot~0}, each run zero-shot from onboard depth and proprioception. Rows follow the order of Section~\ref{sec:exp:real}: reverse-vault, speed-vault, climb-and-step (including the unseen pommel horse), plank bridge, stepping stones, a high platform, and outdoor curbed stairs. The upper rows are whole-body skills, the lower rows the perceptive locomotion that carries the robot between them.}
  \label{fig:real-world}
\end{figure*}

We conduct real-world experiments, indoors and outdoors, to evaluate LightLP under deployment conditions.
The policy tracks a velocity command from onboard depth and proprioception alone.
At deployment, it needs no odometry, no motion-capture data, and no external mapping.
Across these experiments it completes traversals across both structured indoor obstacles and unstructured outdoor terrain.
These deployments demonstrate fully autonomous perceptive whole-body locomotion outdoors on a humanoid, with the policy switching between locomotion and whole-body contact-rich skills from onboard sensing alone.
Fig.~\ref{fig:real-world} shows the deployments, and we introduce each capability in turn.

\emph{reverse-vault.}
The robot executes a dynamic parkour maneuver by jumping toward an obstacle, performing a $360^\circ$ aerial rotation, and using pelvis contact with the obstacle to redirect its motion before landing.
The accumulated angular momentum during landing is then regulated through a second $360^\circ$ rotation, allowing the robot to dissipate impact energy and achieve a stable final pose.
This skill requires a rapid takeoff impulse, accurate obstacle-relative state estimation under temporary visual occlusion, and coordinated whole-body control for contact timing, momentum transfer, and impact absorption.

\emph{speed-vault.}
The speed-vault skill is a high-speed whole-body parkour maneuver that requires rapid momentum generation, transient single-arm support, and whole-body stabilization during flight and landing.
Starting from a running approach, the robot reaches a peak velocity of $3.14$~m/s, plants one hand on the obstacle, redirects its forward momentum, and enters a ballistic aerial phase.
After hand release, the robot regulates its body orientation under inertial loading to reach a landing configuration.
The maneuver requires coordination between unilateral arm support and bilateral leg swing, while the forward kinetic energy before takeoff and the landing impulse must be absorbed without destabilizing the robot.

\emph{Climb-and-step.}
The robot establishes chest-height hand contacts with the obstacle and uses bilateral arm support to lift its body beyond the limits of leg-only actuation.
It then extends its legs, transitions through a kneeling posture on the platform, and recovers to a standing pose before jumping down.
This maneuver challenges the robot to overcome lower-limb torque limitations through sustained upper-body loading and precise whole-body coordination across multiple contact phases.
Furthermore, the drop from the elevated platform requires impact recovery under limited control authority during the aerial phase, so the robot must dissipate the vertical landing impulse at touchdown and regain a stable posture.
We further replace the box with a $50$~cm pommel-horse-like trapezoidal obstacle never seen in training, and the same skill still succeeds on the sloped, non-box profile, showing that it keys on the perceived terrain geometry rather than memorizing a fixed obstacle shape.

\emph{Plank bridge.}
The robot traverses a narrow plank bridge while maintaining both feet on a highly constrained support surface, where even small lateral deviations can result in a loss of contact.
This task requires precise foot placement, perception-aware locomotion, and dynamic balance control under narrow lateral margins.
Unlike walking on flat terrain, the robot must continuously regulate its body motion and foot trajectories because little recovery space remains after a lateral deviation.

\emph{Stepping stones.}
The robot traverses discrete stepping stones with gaps in between, where successful locomotion requires selecting and reaching valid footholds rather than relying on continuous terrain support.
Since the onboard camera provides limited visibility of the foot-placement region, the robot must maintain a memory of previously observed terrain geometry and use this spatial representation to guide future footsteps.
This maneuver evaluates perception-driven foothold selection and memory-based locomotion under partial observability.

\emph{High platform.}
The robot mounts a $30$~cm high platform, representing an extreme step height of approximately $33\%$ of its standing height.
To complete the maneuver, the robot must sustain a prolonged single-leg stance while lifting the swing leg and placing it onto the elevated surface with a large range of motion.
The task presents a significant balance challenge due to the highly asymmetric posture and limited support during the stepping phase, requiring precise foot placement and robust whole-body stabilization.

\emph{Outdoor curbed stairs.}
The robot ascends and descends real outdoor curbed stairs and traverses natural terrains such as grass slopes in a zero-shot setting.
Unlike regular stairs with predictable footholds, curbed stairs require accurate perception of step edges and proactive foot-trajectory adjustment; insufficient foot clearance can cause the swing leg to catch on the curb and result in failure.
This experiment demonstrates the robustness of perceptive locomotion in real-world environments, where the policy must handle depth noise, terrain uncertainty, and previously unseen geometric variations.

\begin{figure*}[t]
  \centering
  \includegraphics[width=\textwidth]{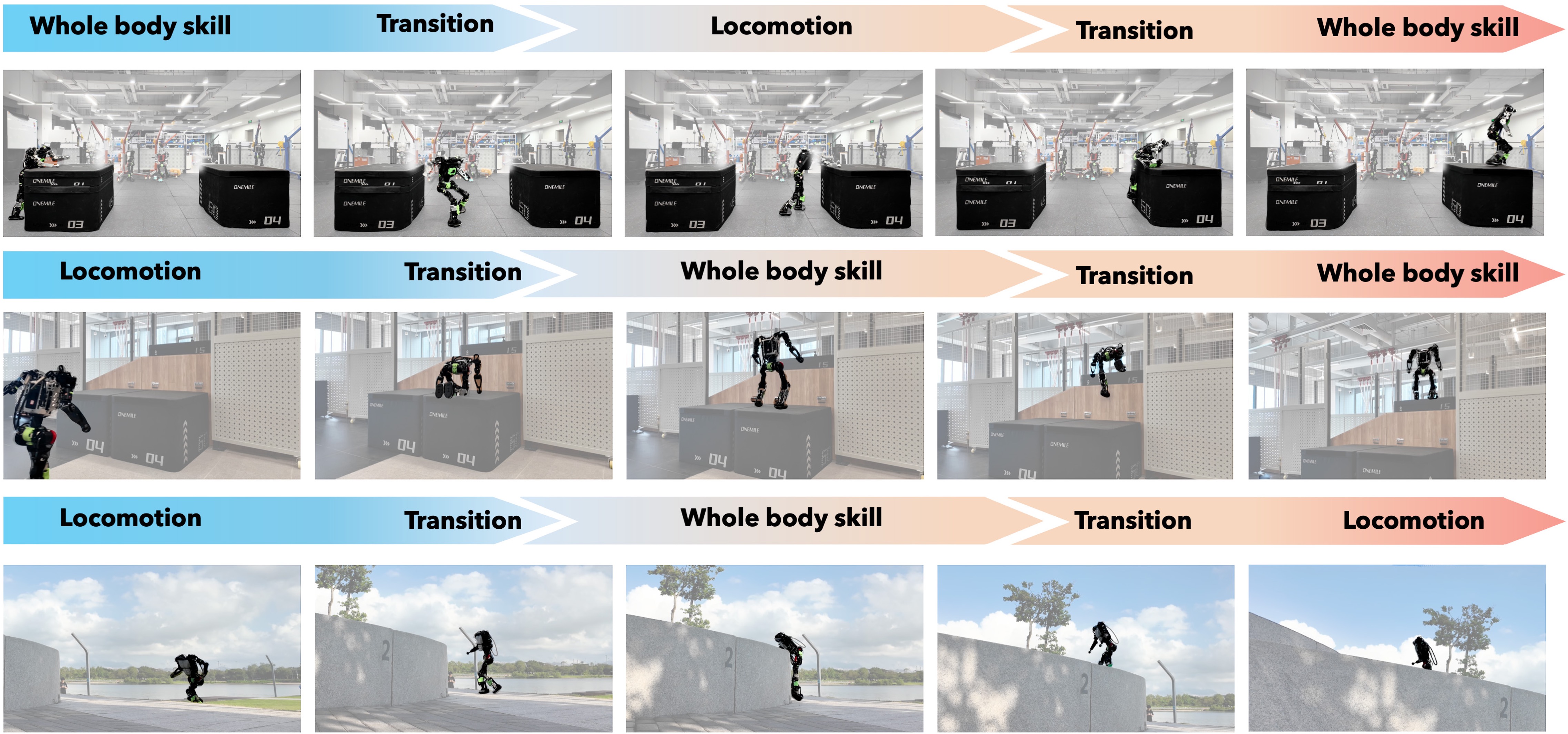}
  \caption{%
    \textbf{Real-world transition courses.} The policy switches between locomotion and whole-body skills on its own, with no external cue. Top: indoors, a climb-and-step followed by a staircase. Bottom: outdoors, crossing a grass field, ascending a staircase, and climbing a higher obstacle.}
  \label{fig:transition-exp}
\end{figure*}

\emph{Autonomous transitions.}
Finally, the policy chains locomotion and skills into continuous courses, deciding on its own and with no external cue when to switch (Fig.~\ref{fig:transition-exp}).
This setting integrates the preceding capabilities, and we analyze it in detail in Section~\ref{sec:exp:real:transition}.

\subsection{Simulation and Benchmark}
\label{sec:exp:sim:skill}
We now turn to simulation, where we benchmark against prior work and ablate each component under controlled, repeatable conditions.
We first evaluate the whole-body skills, namely reference generation and skill learning, and then the perceptive locomotion.

\subsubsection{Reference Generation}
\label{sec:exp:reference}
Learning a \emph{smooth} whole-body skill hinges on obtaining a good reference.
Without one, the robot adopts strange poses while interacting with the object, because an obstacle-misaligned reference collides with the scene.
Starting from the same seed motion, we compare our references against the two widely-used retargeting methods, GMR~\citep{joao2025gmr} and OmniRetarget~\citep{yang2025omniretarget}, with the results reported in Table~\ref{tab:reference-compare}.

\emph{No penetration}: the body must not pass through the obstacle. A direct GMR retarget ignores the obstacle entirely, so the hands and knees penetrate its surface.
\emph{No floating}: the body must actually touch the obstacle rather than hover above it. On climb-and-step, OmniRetarget avoids penetration only by keeping the body just off the surface, so the hands and feet never make contact.
\emph{Object-paired}: the motion is captured together with the obstacle it interacts with.
\emph{Dynamically feasible}: the motion is consistent with the robot's dynamics and actuation limits, with no teleporting jumps or other physically implausible transitions.
As Table~\ref{tab:reference-compare} shows, only our references satisfy all four requirements at once: no penetration, no floating, object-paired, and dynamically feasible.
OmniRetarget sometimes reports a larger nominal range, especially for vaulting, but this is a kinematic overestimate rather than an executable skill.
On speed-vault the failure is sharper: to keep the body clear of the box, the retarget snaps it across in a discontinuous jump and teleports through the obstacle instead of clearing it.
No torque-limited humanoid can execute such a reference, so it is unusable for policy learning.
On climb-and-step, the retargeted motion instead hovers without load-bearing contact, again leaving no continuous behavior to learn.
The contrast with climb-and-step is instructive, since that skill depends directly on precise hand support and therefore exposes the missing contact detail.

\begin{table}[!ht]
\caption{Comparison of whole-body reference-generation methods.}
\label{tab:reference-compare}
\centering
\renewcommand{\arraystretch}{1.3}
\setlength{\tabcolsep}{4pt}
\begin{tabular}{@{}l c c c c c@{}}
\hline
Method & Range & Paired & No pen. & No float. & Feasible \\
\hline
\multicolumn{6}{@{}l}{\emph{Climb-and-step}}\\
GMR~\citep{joao2025gmr}                    & single        & \xmark & \xmark & \xmark & \xmark \\
OmniRetarget~\citep{yang2025omniretarget}  & $50$--$70$~cm & \cmark & \cmark & \xmark & \xmark \\
\textbf{Ours}                              & $50$--$75$~cm & \cmark & \cmark & \cmark & \cmark \\
\hline
\multicolumn{6}{@{}l}{\emph{speed-vault}}\\
GMR                                        & single        & \xmark & \xmark & \xmark & \xmark \\
OmniRetarget                               & $40$--$55$~cm & \xmark & \xmark & \xmark & \xmark \\
\textbf{Ours}                              & $40$--$50$~cm & \cmark & \cmark & \cmark & \cmark \\
\hline
\multicolumn{6}{@{}l}{\emph{reverse-vault}}\\
GMR                                        & single        & \xmark & \xmark & \xmark & \xmark \\
OmniRetarget                               & $40$--$55$~cm & \cmark & \cmark & \xmark & \xmark \\
\textbf{Ours}                              & $40$--$50$~cm & \cmark & \cmark & \cmark & \cmark \\
\hline
\end{tabular}
\end{table}

\subsubsection{Ablation Study}
\label{sec:exp:real:skill}
\begin{figure}[!t]
  \centering
  \includegraphics[width=\columnwidth]{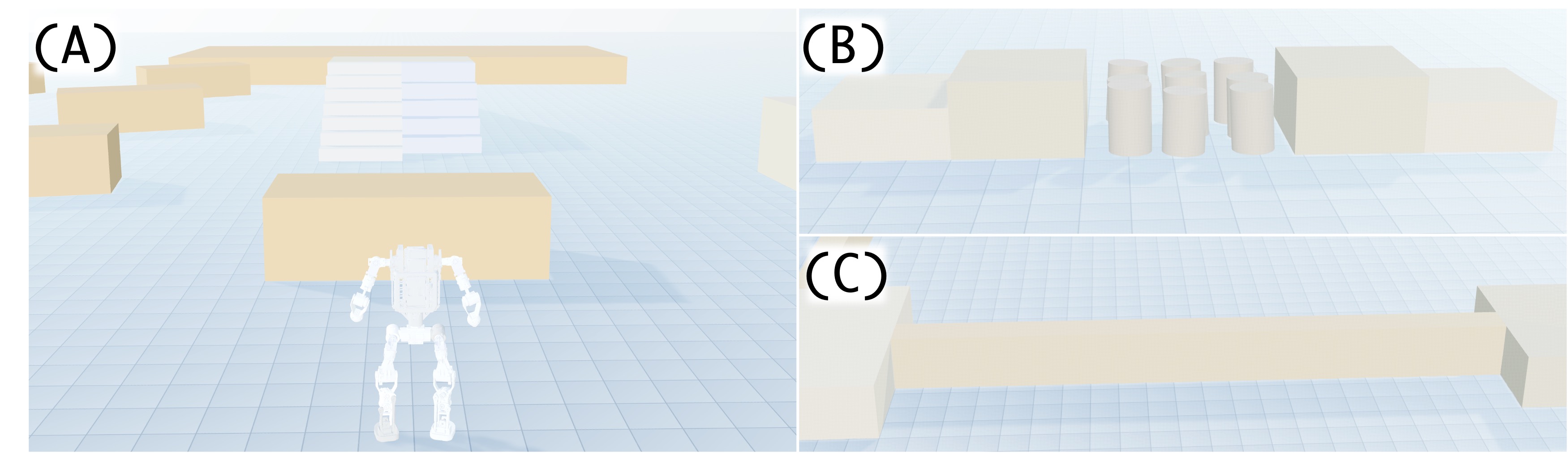}
  \caption{Representative simulation evaluation scenes for whole-body parkour and perceptive locomotion, including obstacle skills, high stairs, stepping stones, and a balance beam.}
  \label{fig:simulation-scenes}
\end{figure}

Table~\ref{tab:real-skill} reports a set of ablation experiments across all skills and perceptive-locomotion terrains.
We evaluate \emph{Ours}, \emph{Ours w/o FT}, \emph{Ours w/o GRU}, \emph{Teacher}, \emph{Teacher w/o ED}, and \emph{Teacher w/o RA}.
The student-side ablations isolate the final fine-tune and recurrent memory, while \emph{Teacher w/o ED} tests whether the expert whole-body skill can be learned without expert distillation and \emph{Teacher w/o RA} tests the effect of reference augmentation.
The simulation evaluation covers both whole-body parkour skills and perceptive locomotion, with representative scenes shown in Fig.~\ref{fig:simulation-scenes}.
In the whole-body-skill rows, each policy is tested across obstacle heights; within each height, the box width and depth are randomized as $x\in[0.7,1.0]$~m and $y\in[0.5,1.5]$~m.
Each reported ablation success rate in Table~\ref{tab:real-skill} is computed over $500$ randomized trials under the corresponding setting.
To keep the comparison controlled, every variant is trained independently under the same environment and a $15{,}000$-iteration budget, so that differences in success rate reflect the removed component rather than training compute.
The depth policy is distilled for $14{,}000$ iterations and then fine-tuned for a further $1{,}000$; the \emph{w/o fine-tune} variant stops after distillation.

\begin{table*}[!t]
\caption{Ablation and benchmark experiments.}
\label{tab:real-skill}
\centering
\renewcommand{\arraystretch}{1.2}
\scriptsize
\setlength{\tabcolsep}{1.5pt}
\resizebox{0.92\textwidth}{!}{%
\begin{tabular}{l c c c c c c c c c c}
\hline
\noalign{\vskip 1pt}
\tablehead{Configuration} & \tablehead{\textbf{Ours}} & \tablehead{\textbf{Ours}\\w/o FT} & \tablehead{\textbf{Ours}\\w/o GRU} & \tablehead{Teacher} & \tablehead{Teacher\\w/o ED} & \tablehead{Teacher\\w/o RA} & \tablehead{PHP\\\citep{wu2026perceptive}} & \tablehead{MGMT\\\citep{zhang2026learning}} & \tablehead{BeamDojo\\\citep{wang2025beamdojo}} & \tablehead{CReF\\\citep{hao2026cref}} \\[1pt]
\hline
\multicolumn{11}{@{}l}{\emph{Whole-body skills: success rate (\%)}}\\
climb-and-step 60 cm / $0.66H$  & $99.2$ & $88.4$ & $54.0$ & $99.9$ & \xmark & $99.9$ & $95$ & $96.2$ & / & / \\
climb-and-step 65 cm / $0.72H$  & $98.8$ & $89.8$ & $56.6$ & $99.2$ & \xmark & \xmark & / & / & / & / \\
climb-and-step 70 cm / $0.77H$  & $90.0$ & $76.4$ & $34.2$ & $99.2$ & \xmark & \xmark & / & / & / & / \\
climb-and-step 75 cm / $0.83H$  & $33.4$ & $17.0$ & $0.0$ & $98.6$ & \xmark & \xmark & / & / & / & / \\
\hline
reverse-vault 40 cm / $0.44H$ & $99.6$ & $74.4$ & $26.4$ & $99.9$ & \xmark & $99.9$ & / & / & / & / \\
reverse-vault 45 cm / $0.50H$ & $99.2$ & $75.4$ & $26.8$ & $99.9$ & \xmark & \xmark & / & / & / & / \\
reverse-vault 50 cm / $0.55H$ & $96.8$ & $72.8$ & $25.4$ & $99.4$ & \xmark & \xmark & / & / & / & / \\
\hline
speed-vault 40 cm / $0.44H$ & $99.0$ & $65.2$ & $23.4$ & $99.9$ & \xmark & $91$ & $99$ & / & / & / \\
speed-vault 45 cm / $0.50H$ & $95.0$ & $68.4$ & $22.6$ & $99.8$ & \xmark & \xmark & / & / & / & / \\
speed-vault 50 cm / $0.55H$ & $93.4$ & $61.8$ & $23.2$ & $99.6$ & \xmark & \xmark & / & / & / & / \\
\hline
\multicolumn{11}{@{}l}{\emph{Perceptive locomotion: success rate (\%)}}\\
Stepping stones        & $99.9$ & $34.6$ & $0$ & $99.9$ & / & / & / & / & $91.7$ & / \\
Balance beam           & $99.9$ & $99.9$ & $53.8$ & $99.9$ & / & / & / & / & $94.3$ & / \\
Stairs (low)           & $99.9$ & $99.9$ & $47.6$ & $99.9$ & / & / & / & $99.9$ & / & / \\
Stairs (middle)        & $99.9$ & $89.4$ & $19.0$ & $99.9$ & / & / & $99.9$ & $99.9$ & / & $97.9$ \\
Stairs (high)          & $83.4$ & $80.8$ & $12.4$ & $83.0$ & / & / & / & $98.7$ & / & / \\
Stones ($+$height)     & $95.6$ & $24.4$ & $0$ & $98.6$ & / & / & / & / & / & / \\
Beam ($+$yaw)          & $99.9$ & $99.9$ & $42.2$ & $99.9$ & / & / & / & / & / & / \\
\end{tabular}}
\vspace{1pt}
\rule{0.92\textwidth}{0.4pt}
\parbox{0.92\textwidth}{\footnotesize Skill rows use randomized boxes with $x\in[0.7,1.0]$~m and $y\in[0.5,1.5]$~m. / indicates settings not reported or outside the method's scope; \xmark indicates failure to converge.}
\end{table*}

The most pronounced ablation effect comes from the final fine-tune.
Distillation from the expert already gives the depth policy a usable version of each behavior, but the policy still suffers from residual errors after replacing the teacher's height scan with onboard depth.
This drop is especially clear on foothold-sensitive tasks such as stepping stones, where small perception or timing errors directly lead to missed contacts.
The final RL fine-tune largely closes this gap, bringing the depth policy back toward teacher-level performance by optimizing task success under the student's own observations.

We also compare feed-forward MLP and recurrent GRU architectures.
The non-recurrent policy suffers a cliff-like drop in success, because the teacher observes privileged information such as the body's local linear velocity whereas the deployable student does not.
A memory-based network can partially recover this missing state in its latent representation by integrating recent proprioception and depth observations; we observed the same conclusion when replacing the GRU with an LSTM.
Comparing \emph{Ours} with the \emph{Teacher} also shows the limit of memory: it mitigates the missing velocity and height-scan information, but cannot fully replace them.
For example, in the $75$~cm climb-and-step setting, the camera motion makes the obstacle height difficult to infer reliably during the approach, leading to a sharp drop in success.

The teacher-side ablations ask two related questions.
\emph{Teacher w/o ED} tests whether an expert whole-body skill can be learned directly from sparse reward without an expert-distillation stage, while \emph{Teacher w/o RA} tests whether a single reference trajectory can provide generalization across obstacle variations.
Together, they point to the difficulty of learning high-dynamic whole-body skills from sparse reward and a single seed motion alone.
Following the HIL-style setting and removing expert distillation, the teacher fails to converge even though it still has privileged observations.
The reason is not merely observation quality: to learn whether and how to traverse an obstacle, the robot must discover a sequence of forceful contacts with the object, and the exploration space becomes too large for reward-only learning.
The policy often collides with the obstacle and settles into a poor local minimum instead of finding the coordinated contact sequence needed for vaulting or climbing.
Using only a single reference without augmentation also overfits the teacher to that one trajectory and obstacle height.
It may succeed at the seed configuration, but fails to cover the range in Table~\ref{tab:real-skill}; even on speed-vault, the success drops, suggesting that multiple terrain-paired references provide not only height coverage but also a broader exploration space for refining contact timing and placement.

Beyond scale, we test generalization to obstacle \emph{shape}.
For reverse-vault and speed-vault we replace the training box with a pommel-horse-like trapezoidal obstacle never seen in training.
Table~\ref{tab:real-skill-shape} shows that both skills transfer to the new shape, confirming that they key on the terrain the robot perceives rather than on a fixed obstacle profile.

\begin{table}[t]
\caption{Success rate (\%) on a trained box vs.\ an unseen pommel-horse (trapezoidal) obstacle.}
\label{tab:real-skill-shape}
\centering
\renewcommand{\arraystretch}{1.2}
\setlength{\tabcolsep}{5pt}
\begin{tabular}{@{}l c c c c@{}}
\hline
 & \multicolumn{2}{c}{reverse-vault} & \multicolumn{2}{c}{speed-vault} \\
\cline{2-3}\cline{4-5}
Method & Box & Horse & Box & Horse \\
\hline
\textbf{Ours (Student)} & $99.9$ & $93.4$ & $99.9$ & $95.3$ \\
\hline
\end{tabular}
\end{table}

\subsubsection{Benchmark}
\label{sec:exp:sim:loco}
We evaluate our student policy against state-of-the-art methods and, on the same terrains, ablate the components that produce it.
We compare with four published humanoid controllers as policy-level configurations: motion generation plus motion tracking (MGMT)~\citep{zhang2026learning}, PHP~\citep{wu2026perceptive}, BeamDojo~\citep{wang2025beamdojo} for sparse footholds, and CReF~\citep{hao2026cref} for stairs.
These methods cover complementary terrains: MGMT and PHP focus on reference-motion whole-body locomotion, BeamDojo on sparse footholds, and CReF on stairs. Our student policy is evaluated across all of them, including whole-body skills such as climb-and-step and speed-vault, stepping stones with varying height, and balance beams with yawed approaches.
In Table~\ref{tab:real-skill}, ``/'' indicates that the method does not report that experiment.

\emph{Normalized obstacle height.}
To compare robots of different sizes, we report each obstacle height as a ratio $h/H$, where $h$ is the obstacle height and $H$ is the robot's standing height.
This ratio gives a simple scale-normalized measure of how large the obstacle is relative to the robot.
Under this metric, PHP reports whole-body obstacle traversal mainly around $0.45H$ and $0.60H$, and does not demonstrate the higher normalized heights tested here.
MGMT similarly reports whole-body climbing around $0.60H$.
By contrast, our student policy is evaluated up to $0.83H$ for climb-and-step and shows the same advantage on speed-vault, while maintaining higher success at the overlapping settings.
This indicates stronger generalization across obstacle scales and pushes the behavior closer to the robot's physical limits.
We attribute this to the high-precision, dynamically feasible references produced by our method; combined with the final task-reward fine-tune, these references allow the student policy to retain high success even at larger normalized obstacle heights.

\emph{Coverage across terrains.}
Beyond whole-body skills, Table~\ref{tab:real-skill} also compares perceptive locomotion settings.
BeamDojo is designed specifically for sparse footholds such as stepping stones and balance beams, yet its reported success remains lower than ours on the overlapping settings.
For stairs, we divide the task into low ($15$~cm), middle ($25$~cm), and high ($35$~cm) levels; CReF reaches only the middle level and still reports lower success than our student policy.
We attribute this broader terrain coverage to the reward design used during training and to the final fine-tune, which optimizes task success under the deployable depth observations.
We further test stepping stones with height variation and balance beams with yawed approaches, where the student policy still maintains high success.
Taken together, these results show that the proposed pipeline realizes a broad set of perceptive locomotion and whole-body parkour skills within a single deployable policy.
This moves humanoid whole-body control beyond isolated terrain-specific behaviors toward a unified controller that can coordinate stepping, balancing, climbing, and vaulting within one learned policy.
\subsection{Transitions}
\label{sec:exp:real:transition}
Fig.~\ref{fig:transition-exp} demonstrates the transition courses in the real world, covering skill-to-locomotion, skill-to-skill, and locomotion-to-skill switches.
These three settings jointly verify that the transition mechanism is effective across the full set of behavior changes required by the proposed pipeline.
The decision is made from the policy's depth observation together with the velocity command: in the transition group, the policy effectively learns the composition of skills and selects, at each moment, which behavior is needed to follow the command through the current terrain.

\subsubsection{Ablation Study}
\begin{table}[t]
\caption{Transition ablation and command-adherence test: transition-and-skill success rate (\%).}
\label{tab:transition-ablation}
\centering
\renewcommand{\arraystretch}{1.2}
\begin{tabular}{@{}l c@{}}
\hline
Setting & Result \\
\hline
\multicolumn{2}{@{}l}{\emph{Transition-group ablation}}\\
Plane locomotion $+$ skills, w/o transition group          & $0$ \\
Rough perceptive locomotion $+$ skills, w/o transition group & $33$ \\
w/o AMP                                                    & $51$ \\
\textbf{Ours (with transition group)}                      & $98$ \\
\hline
\multicolumn{2}{@{}l}{\emph{Command adherence near an obstacle}}\\
Reverse-away command in front of obstacle                  & True \\
\hline
\end{tabular}
\end{table}

\begin{figure*}[!t]
  \centering
  \includegraphics[width=\textwidth]{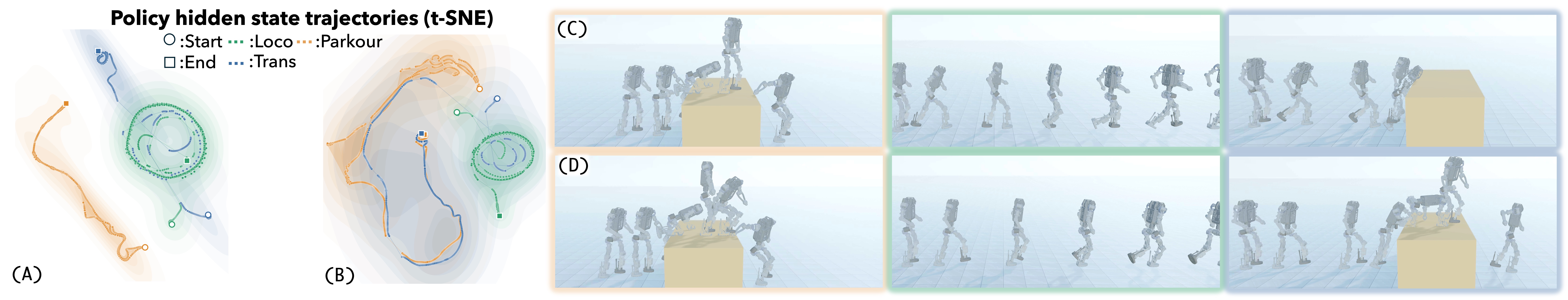}
  \caption{%
    t-SNE visualization of the RNN hidden states during transition courses, colored by behavior type: flat locomotion, transition, and parkour; rollout snapshots are paired with the corresponding embeddings.
    \textbf{(A,C)} Without the transition group, the hidden state stays confined to the perceptive-locomotion region even when the terrain ahead demands a skill, so the policy keeps walking or steps in place and never enters skill mode.
    \textbf{(B,D)} With the transition group, the hidden state leaves the locomotion region and moves into the parkour-skill region as the terrain requires, while the transition states form a bridge between the two clusters.
    The paired contrast shows that the recurrent policy learns to encode which mode to enter from the observed terrain and velocity command alone.}
  \label{fig:tsne-transition}
\end{figure*}

We count a trial as successful only when the policy both switches to the behavior required by the terrain and completes the resulting skill or locomotion segment.
Evaluated over $100$ randomized trials per setting, Table~\ref{tab:transition-ablation} shows that the transition group is essential for this end-to-end behavior.
Without it, a policy trained only on flat locomotion and isolated whole-body skills never learns how to connect them, and the success rate remains $0\%$.
Even when rough-terrain perceptive locomotion is added, removing the transition group leaves the policy without a learned composition strategy: in front of an obstacle, it often continues the locomotion behavior, collides with the wall, or repeatedly steps in place instead of invoking the required whole-body skill.
This raises success only to $33\%$.
With the transition group, the policy identifies from depth and the velocity command which skill is needed in the current scene and completes the obstacle traversal, raising success to $98\%$.
To examine how this switch is represented internally, we visualize the RNN hidden states with t-SNE in Fig.~\ref{fig:tsne-transition}.
With the transition group, the hidden state can leave the perceptive-locomotion region and enter the parkour-skill region when the terrain requires a skill.
Without the transition group, the hidden state remains in the perceptive-locomotion region, explaining why the policy keeps walking or stepping in place instead of executing the required transition.

We next ablate the adversarial motion prior (AMP), which is needed not only for switching but also for deployable motion quality.
Without AMP, the policy can still react to terrain cues, but the unconstrained objective produces unnatural motions and awkward contacts that are difficult to transfer directly to hardware.
With AMP, the policy remains close to the learned skill manifold, yielding both a reliable transition and a physically plausible whole-body motion (Table~\ref{tab:transition-ablation}).

We further probe whether the policy truly acts on the command rather than being drawn to any obstacle in view.
This failure can arise in reference-bound whole-body policies~\citep{wu2026perceptive}: because both locomotion and obstacle traversal are learned from a limited set of reference motions, the policy may not become robust to velocity commands that contradict the reference, and an observed obstacle can trigger forward traversal even when the command asks the robot to retreat.
Standing in front of an obstacle, we therefore command the robot to reverse away from it.
Our policy retreats as commanded (Table~\ref{tab:transition-ablation}), confirming that its behavior is decided jointly by the velocity command and the depth observation rather than by the mere presence of an obstacle.

\section{Conclusion}
\label{sec:conclusion}

We presented \texttt{Light-Loco-Parkour}, an end-to-end perceptive whole-body locomotion system for a humanoid robot.
The central result is a single deployable policy that reads onboard depth and a velocity command, then selects and executes the appropriate behavior across both perceptive locomotion and whole-body parkour skills, without a reference input, skill label, hand-coded gate, or runtime motion graph.
This capability is enabled by three components: high-precision reference--object pairs that align whole-body motion with the terrain contacts required for each skill, RL fine-tuning that turns each skill into a terrain-conditioned behavior, and transition-group training that lets switching between locomotion and skills emerge from sparse reward.
Deployed zero-shot from simulation onto Lightbot~0, the policy traverses climb-and-step, reverse-vault, and speed-vault obstacles in both indoor and outdoor environments, showing that whole-body contact and perceptive locomotion can be unified within one learned controller.

Several limitations are worth stating plainly.
The seed-collection step of our data pipeline still requires a human-in-the-loop to align each video with a candidate obstacle, and we expect that this manual step is a real ceiling on how fast the skill set can be expanded.
The transition policy currently couples a small, discrete set of skills, and we observe degraded behavior when obstacles overlap or appear in close succession, a problem the current sparse-reward setup does not fully resolve.
Finally, the fixed chest-mounted depth camera limits both whole-body manipulation of the scene and the perceptive coverage available to the locomotion controller, particularly when the upper body occludes the camera mid-skill.

These limitations suggest the directions we are most interested in next.
We plan to scale the pipeline to much larger reference sets and study whether richer transition behavior emerges as data grows, rather than having to be hand-shaped.
We also believe the perceptual bottleneck can be addressed through active perception (camera or head gaze controlled by the policy itself) or by giving the policy an explicit short-term memory of the scene, so that mid-skill occlusion does not erase what the robot already saw on approach.

\section*{Acknowledgment}
The authors would like to thank colleagues at Light Origins for helpful discussions and hardware support.

\bibliographystyle{IEEEtranN}
\bibliography{references}

\end{document}